\documentclass[a4paper, 10pt, conference]{ieeeconf}      
\usepackage{FG2027}

\FGfinalcopy 

\IEEEoverridecommandlockouts                              
\usepackage{amssymb}  

\usepackage{algorithm}
\usepackage{algpseudocode}
\usepackage{graphicx}
\usepackage{subcaption}
\usepackage{multirow}
\usepackage{amsmath}
\usepackage{booktabs} 
\usepackage{hyperref}
\usepackage[table]{xcolor}

\def\FGPaperID{255} 

\title{\LARGE \bf
Spatio-Temporal Multi-Order Interaction for Video Recognition
}

\author{\parbox{16cm}{\centering
    {\large Kevin D. Nguetche , Lilian Hollard, Eric Granger, Marco Pedersoli}\\
    {\normalsize
    LIVIA, Dept. of Systems Engineering, ETS Montreal, Canada\\
International Laboratory on Learning Systems (ILLS)}}
}

\begin{document}

\ifFGfinal
\thispagestyle{empty}
\pagestyle{empty}
\else
\author{Anonymous FG2027 submission\\ Paper ID \FGPaperID \\}
\pagestyle{plain}
\fi
\maketitle

\begin{abstract}

 Many temporal approaches for video recognition rely on explicit cross-channel aggregation projections, which have quadratic complexity in the number of channels, introducing significant computational overhead. In addition, the spatio-temporal representation used for video recognition is biased either toward local order interactions (for CNNs) or toward global order interactions (for transformers). Starting from the multi-order spatial representation of MogaNet, our method introduces two complementary contributions to tackle these problems. First, for the temporal modelling, we introduce parallel multi-order processing with depthwise-only convolutions and implicit gating-based aggregation, which is linear in the number of channels, inducing a considerable computational reduction. Second, we propose a dual interaction between spatial and temporal dimensions, which allows a better spatio-temporal representation without increasing the computational cost.
  By combining these two contributions, our proposed VideoMogaNet achieves performance on par with or better than state-of-the-art action recognition approaches across multiple video benchmarks (UCF101, HMDB51, Kinetics-400, Diving48), while considerably reducing the cost of the temporal branch. Our code is included in the suppl. materials and will be made public.

\end{abstract}

\section{INTRODUCTION}

\label{sec:intro}

The exponential growth of online video data has positioned video recognition 
as a cornerstone of modern computer vision, powering diverse applications 
ranging from intelligent surveillance~\cite{kay2017k400} to sports analysis~\cite{wu2022survey, xu2024finesports} 
and automated content creation~\cite{huang2023causalainer, xu2024finesports}. Historically, convolutional architectures 
such as I3D~\cite{carreira2017quo} and SlowFast~\cite{feichtenhofer2019slowfast} 
have dominated this field through their local inductive biases, which prove 
particularly effective for learning on limited data. The advent of Vision 
Transformers~\cite{dosovitskiy2021vit} catalyzed the adoption of attention 
mechanisms for video understanding, leading to architectures like 
TimeSformer~\cite{bertasius2021space}, Video Swin 
Transformer~\cite{liu2022video}, and UniformerV2~\cite{li2022uniformerv2}, 
capable of effectively capturing global spatio-temporal dependencies. More 
recently, state space models such as VideoMamba~\cite{park2024videomamba} 
and their successors~\cite{li2024videomamba, lu2025snakes, dao2024mamba2} 
have demonstrated attractive linear complexity for processing long video 
sequences. Despite these notable architectural advances, fundamental 
limitations persist in efficient spatio-temporal modeling.

\begin{figure}[t]
  \centering
  \includegraphics[width=\linewidth]{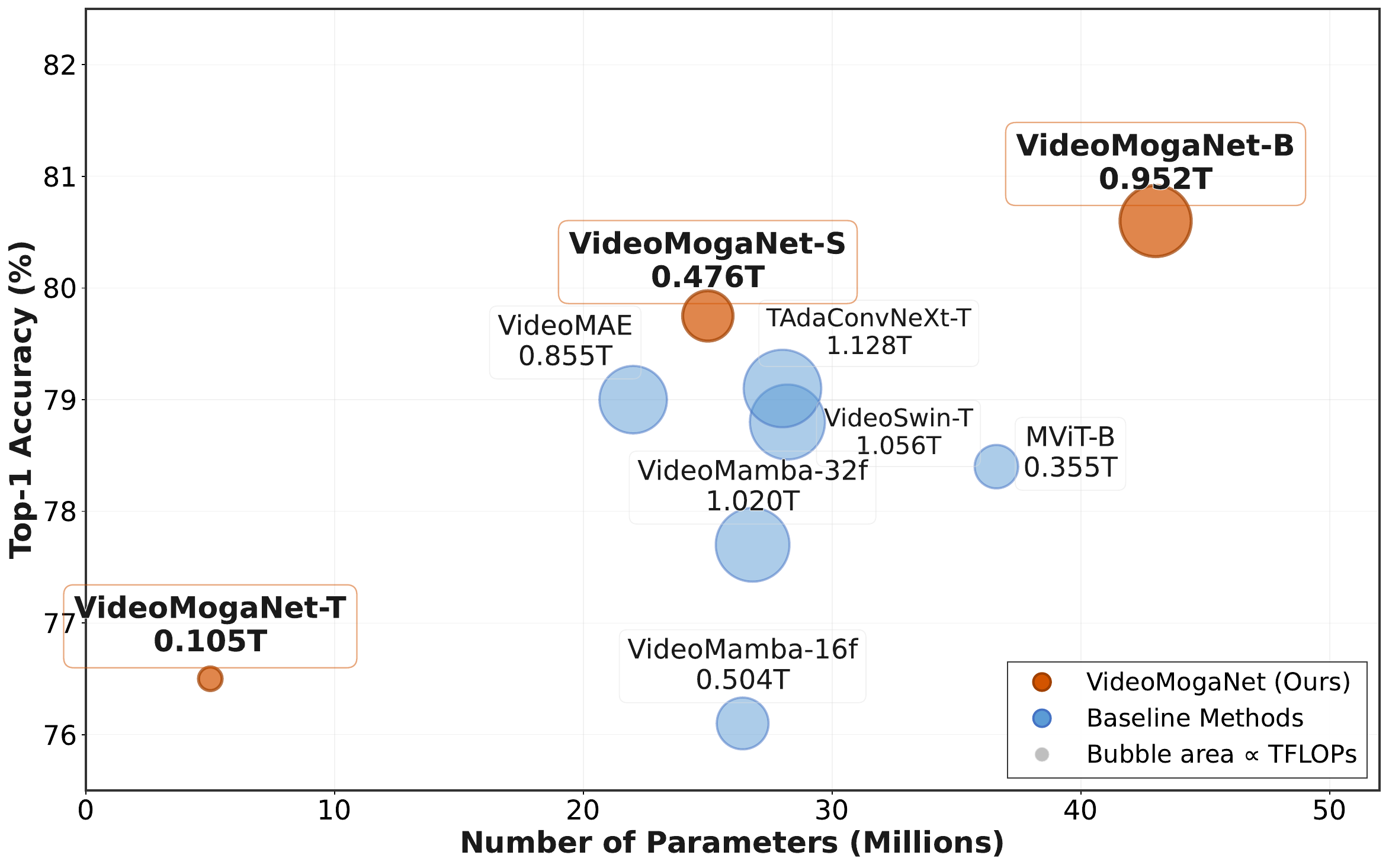}
  \caption{Top-1 accuracy on Kinetics-400 at $224^2$ resolution versus 
  model size (M parameters), with bubble size indicating computational 
  cost (TFLOPs).} 
  \label{fig:teaser}
\end{figure}

Current video architectures face two major challenges. First, Transformers 
suffer from quadratic complexity $\mathcal{O}(T^2 H^2 W^2)$ where \( T \), 
\( H \), and \( W \) denote the temporal length, height, and width of the 
input video, respectively. This makes video processing prohibitively 
expensive: VideoSwin-T~\cite{liu2022video} demands 1.056 TFLOPs for 32 
frames on Kinetics-400, while VideoMAE~\cite{tong2022videomae} requires 
$0.855$ TFLOPs for 16 frames. Although state space models like 
VideoMamba~\cite{park2024videomamba} offer more favorable linear complexity, they still require 32 frames to achieve 77.7\% top-1 
accuracy (at 1.020 TFLOPs) on Kinetics-400, similar to Transformers.

A clear pattern emerges: while Transformers, SSMs, and Modern ConvNets 
can operate with a small number of frames, they rely in practice on long sequences to remain competitive, significantly increasing inference latency 
and memory requirements. In contrast, our approach achieves competitive 
performance from as few as 8 frames, as our temporal modeling encodes interactions of all 
orders while explicitly privileging middle-order ones, which are more 
expressive and discriminative, while remaining temporally scalable for 
long-range modeling. Yet computational efficiency alone does not determine representation quality: a more fundamental question concerns the orders of interactions that video architectures encode.

The \textit{quality} of 
spatio-temporal representations depends critically on the orders of 
interactions that a model encodes. In video, actions unfold across 
multiple spatial scales and temporal extents simultaneously: 
recognizing ''running'' requires not only detecting local textures 
(low-order) or global scene context (high-order), but also structured 
part-whole compositions such as coordinated limb movements spanning 
multiple frames (middle-order). A fundamental architectural question remains: \textbf{what orders 
of interactions} do video models privilege in their representational 
learning? Recent analyses through cooperative game 
theory~\cite{deng2021discovering,ren2024cooperative} formally establish 
that DNNs naturally encode low-order and high-order interactions, but 
suffer from a structural bottleneck for middle-order interactions, a 
phenomenon theoretically proven and empirically verified across multiple 
architectures. In practice, classical ConvNets encode interactions of all 
orders but massively favor \emph{low-order} interactions (local textures, 
edges in adjacent regions), ensuring optimization stability but lacking 
the global semantics required for understanding complex actions. 
Conversely, Transformers also encode interactions of all orders but 
privilege \emph{high-order} interactions (global context at the entire 
scene scale)~\cite{park2023understanding}, at the cost of high 
computational complexity and increased sensitivity to dataset bias and 
rare configurations. Yet, the most discriminative representations for 
video action recognition emerge from \emph{middle-order} interactions, 
namely structured part-whole compositions 
(\textit{e.g.}, arms$+$legs$\rightarrow$running), intermediate spatial 
relationships between distant regions, and actions and gestures spanning 
multiple frames. 
The Representation Bottleneck analysis of Deng~\textit{et al.}~\cite{deng2021discovering}, 
the Shapley Interaction framework of Zhang~\textit{et al.}~\cite{zhang2020interpreting}, 
and MogaNet~\cite{li2022moganet} for images demonstrate that explicitly 
structuring aggregation to encode interactions of all orders while 
privileging middle-order ones significantly improves representation 
quality. The current video literature lacks architectures adopting this 
principle. An effective spatio-temporal architecture should
\textbf{favor and modulate middle-order interactions}. It should also
preserve the local stability of low-order interactions and the global
awareness of high-order interactions, while remaining lightweight.

To address this gap, we introduce \textbf{VideoMogaNet}, a 
spatio-temporal architecture built on two complementary contributions: i) a low-cost temporal modeling module that captures inter-frame dynamics at all interaction orders via parallel depthwise 
channel-split convolutions while explicitly privileging middle-order ones, and ii) a spatio-temporal fusion design introducing 
a dual-level interaction between spatial and temporal dimensions, where 
existing approaches are limited to a single interaction. As illustrated in 
Fig.~\ref{fig:teaser}, VideoMogaNet achieves a favorable trade-off between 
accuracy, computational cost, and temporal efficiency on Kinetics-400.

Our main contributions are as follows:
\begin{itemize}
    \item A multi-order temporal modeling module with lightweight computational cost 
    ({\raise.17ex\hbox{$\scriptstyle\mathtt{\sim}$}}0.282 GFLOPs), 
   representing a significant reduction in temporal cost compared to standard 
    temporal approaches, achieving competitive performance from as few 
    as 8 frames while remaining temporally scalable for long-range 
    modeling.

    \item A dual-level spatio-temporal fusion design, where existing 
    approaches are limited to a single interaction, achieving comparable 
    or superior performance to state-of-the-art approaches.

    \item Extensive validation across multiple benchmarks with distinct 
    characteristics, demonstrating the robustness and generalization of our 
    approach.
\end{itemize}

\begin{figure*}[t]
  \centering
  \begin{minipage}[t]{0.79\linewidth}
    \centering
    \includegraphics[width=\linewidth, height=0.26\textheight, keepaspectratio]{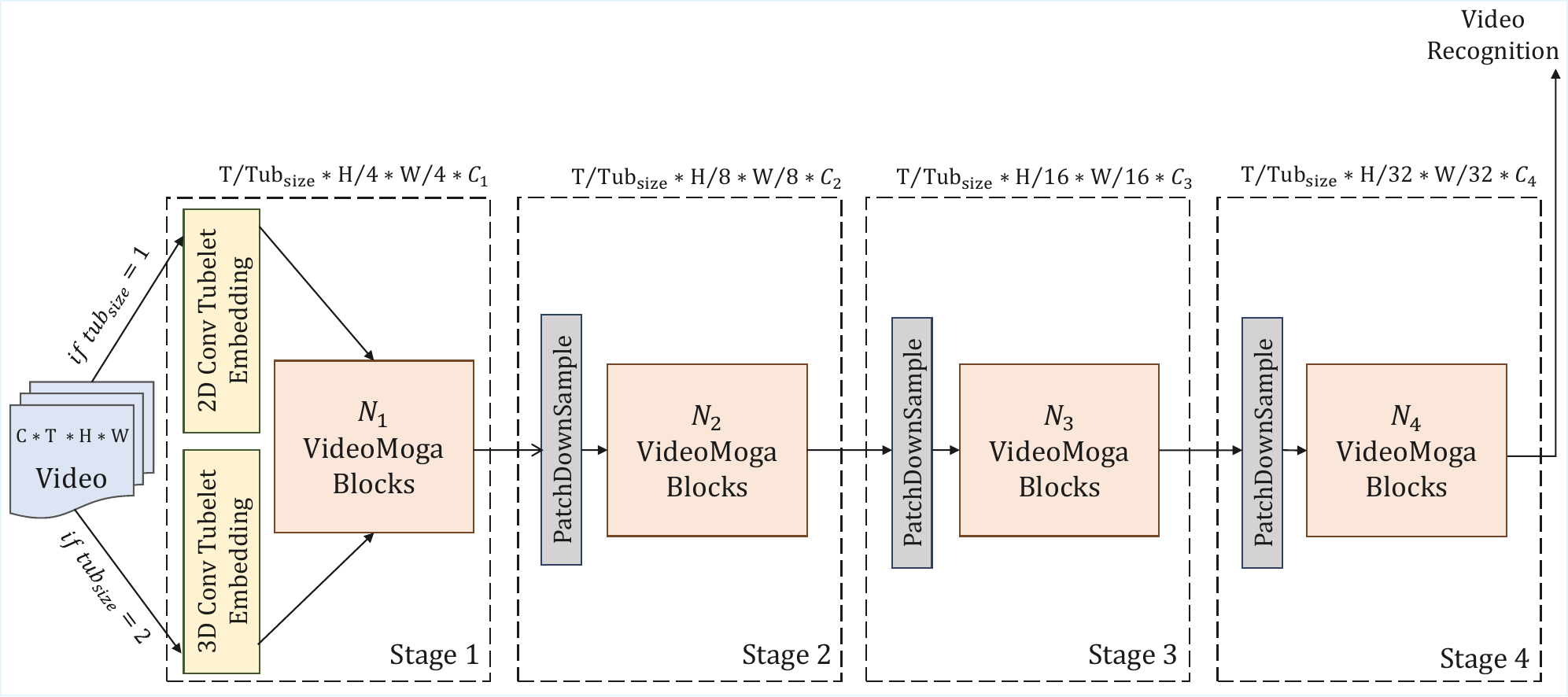}
  \end{minipage}%
  \hfill
  \begin{minipage}[t]{0.19\linewidth}
    \centering
    \includegraphics[width=\linewidth, height=0.26\textheight, keepaspectratio]{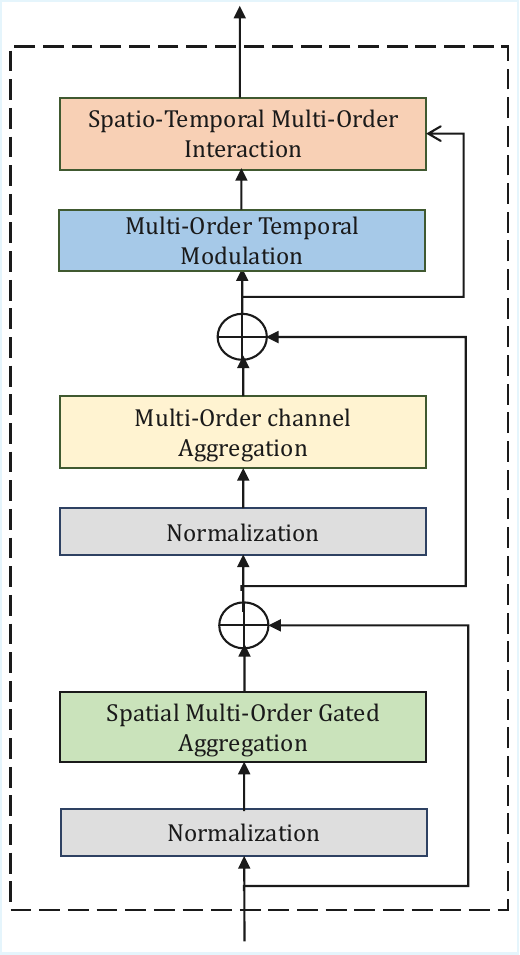}
  \end{minipage}
  \caption{\textbf{VideoMogaNet architecture.} \textit{\textbf{Left}:}
  Four-stage hierarchy following~\cite{liu2022video}; unlike backbones
  that double channels ($C_i{=}2C_{i-1}$), VideoMogaNet adopts a variable
  channel growth ratio across stages. For Base:
  $\{C_i\}{=}\{64, 160, 320, 512\}$,
  $\{N_i\}{=}\{4, 6, 22, 3\}$,
  FFN ratio ${=}\{8, 8, 4, 4\}$, tubelet size ${=}1$.
  \textit{\textbf{Right}:} VideoMoga Block.}
  \label{fig:overall_videomoganet}
\end{figure*}

\section{RELATED WORK}
\label{sec:related_work}

\noindent \textbf{Video Recognition Architectures}.
Convolutional networks have historically dominated video recognition 
through their local inductive biases. 3D architectures such as 
I3D~\cite{carreira2017quo}, C3D~\cite{tran2015learning}, and 
SlowFast~\cite{feichtenhofer2019slowfast} jointly model spatial and 
temporal dimensions, while factorized variants like 
R(2+1)D~\cite{tran2018closer} and X3D~\cite{feichtenhofer2020x3d} reduce 
computational cost. These models encode interactions of all orders but 
massively favor \emph{low-order} interactions (local textures, adjacent 
edges), which limits the capture of long-range dependencies. The advent 
of Vision Transformers~\cite{dosovitskiy2021vit} catalyzed the adoption 
of attention mechanisms for video, leading to 
TimeSformer~\cite{bertasius2021space}, ViViT~\cite{arnab2021vivit}, 
Video Swin Transformer~\cite{liu2022video}, and 
UniformerV2~\cite{li2022uniformerv2}. These models also encode 
interactions of all orders but privilege \emph{high-order} 
interactions~\cite{park2023understanding}, at the cost of quadratic 
complexity $\mathcal{O}(T^2 H^2 W^2)$ and tens of GigaFLOPs for 
temporal encoding. State space models such as 
VideoMamba~\cite{park2024videomamba, li2024videomamba, chen2024video} 
and VideoMambaPro~\cite{lu2025snakes} offer attractive linear complexity, 
yet still require 16 to 32 frames to maintain competitive performance. 
Post-ViT \emph{Modern ConvNets}, such as ConvNeXt~\cite{liu2022convnext}, 
TAdaConv~\cite{huang2021tada}, and MogaNet~\cite{li2022moganet}, combine 
the Transformer macro-architecture with convolutional inductive biases. 
MogaNet demonstrated for images that explicitly structuring aggregation 
to encode interactions of all orders while privileging \emph{middle-order} 
interactions (discriminative part-whole compositions) outperforms 
approaches exclusively favoring extreme orders. However, no video 
architecture proposes to date a lightweight-cost multi-order temporal 
modeling built on this principle.

\noindent \textbf{Efficient Temporal Modeling}.
For video temporal modeling, temporal attention mechanisms 
(TimeSformer~\cite{bertasius2021space}, Video Swin~\cite{liu2022video}) 
require tens of GigaFLOPs, temporal shift and difference operations 
(TSM~\cite{lin2019tsm}, TDN~\cite{wang2020tdn}) are lightweight but 
limited to local interactions, and dynamic convolutions 
(TAdaConv~\cite{huang2021tada}, TAM~\cite{liu2021tam}) calibrate weights 
according to content without explicitly structuring temporal interaction 
orders. Our approach addresses these limitations through two synergistic 
contributions: on the one hand, the MOTM (\textit{Multi-Order Temporal 
Modulation}), a multi-order temporal modeling module with parallel 
channel-split processing, depthwise-only convolutions and implicit 
gating-based aggregation (O(C) cost, ${\sim}$0.282~GFLOPs), representing a significant reduction in computational cost compared to standard temporal approaches; on the other hand, the ST-MOI 
(\textit{Spatio-Temporal Multi-Order Interaction}), a dual-level 
interaction design between spatial and temporal dimensions, where existing 
approaches are limited to a single interaction.

\noindent \textbf{Multi-Order Interactions}.
Cooperative game theory provides a formal framework for analyzing learned 
representations~\cite{deng2021discovering,zhang2020interpreting}. These 
analyses establish that interaction significance grows with network depth, 
reaching its maximum in intermediate representations. \emph{Low-order} 
interactions capture widely-shared local textures, \emph{high-order} 
interactions model global context but tend to memorize rare configurations 
and are prone to overfitting~\cite{cheng2021game}, while \emph{middle-order} 
interactions are primarily responsible for encoding discriminative 
high-level representations~\cite{deng2021discovering}, capturing structured 
compositions (\textit{e.g.}, arms$+$legs$\rightarrow$running). Such 
middle-order interactions thus correspond to coherent semantic structures 
that neither the purely local interactions of ConvNets nor the global 
context of Transformers can effectively capture. Empirical observations 
show that classical ConvNets encode all orders but massively favor 
low-order interactions, while Transformers, which also encode interactions 
of all orders, nevertheless privilege \emph{high-order} 
ones~\cite{park2023understanding}, leaving middle-order interactions 
systematically underexploited. MogaNet~\cite{li2022moganet} demonstrates 
for images that explicitly structuring aggregation to encode interactions 
of all orders while privileging \emph{middle-order} ones significantly 
improves representation quality, a property we apply for the first time 
to temporal video modeling.

\section{PROPOSED METHOD: VIDEOMOGANET}
\label{sec:method}

\begin{figure*}[t]
  \centering
  \begin{minipage}[t]{0.58\linewidth}
    \centering
    \includegraphics[width=\linewidth, height=0.25\textheight, keepaspectratio]{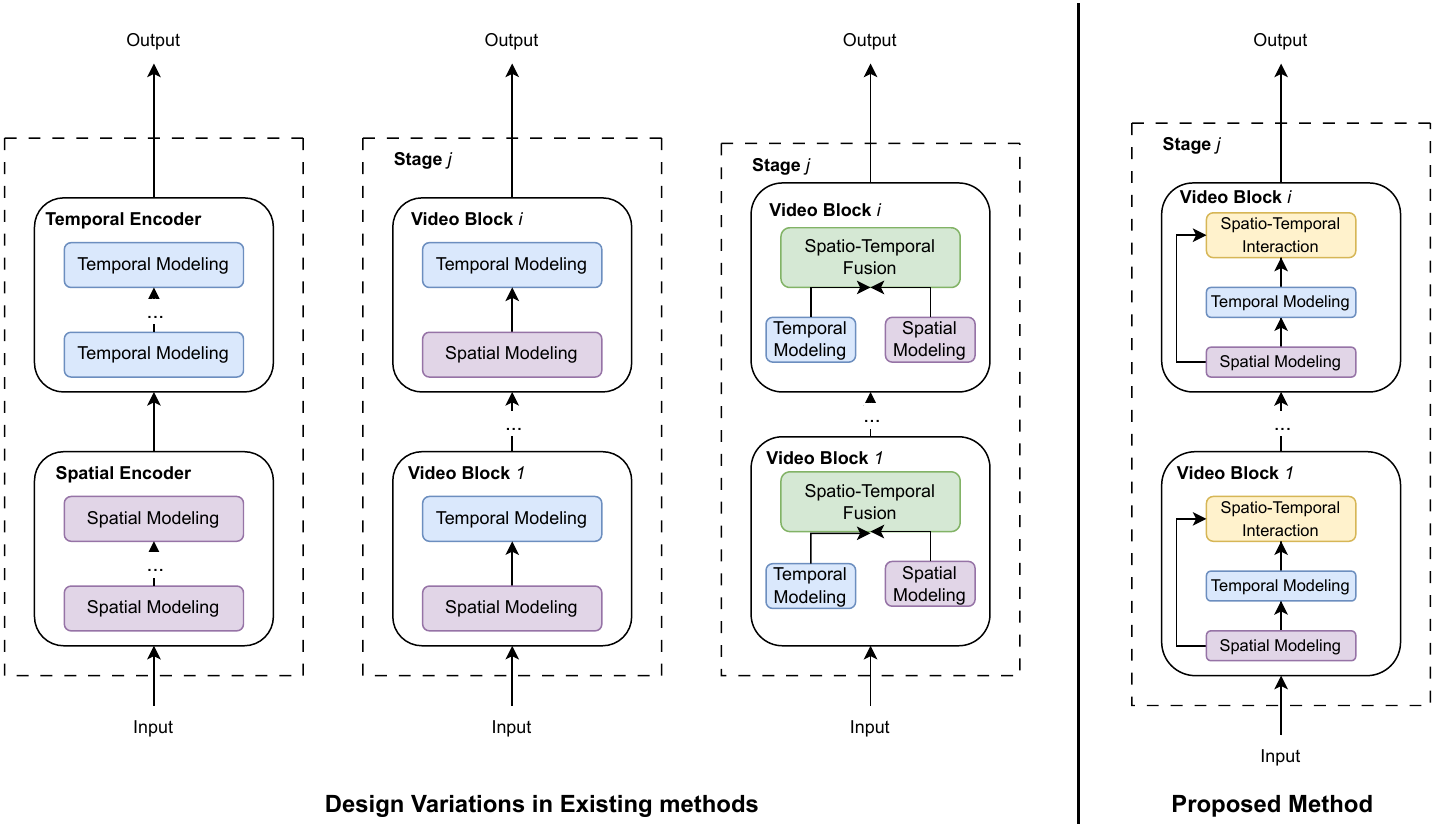}
  \end{minipage}%
  \hfill
  \begin{minipage}[t]{0.40\linewidth}
    \centering
    \includegraphics[width=\linewidth, height=0.25\textheight, keepaspectratio]{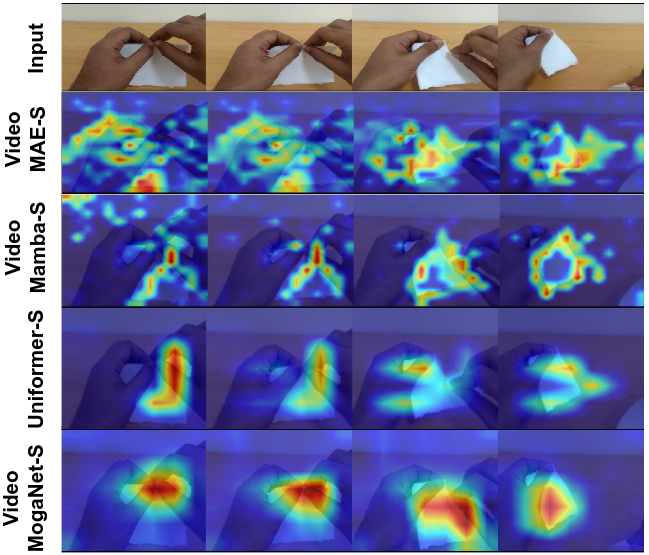}
  \end{minipage}
  \caption{ \textit{\textbf{Left}:} Spatio-temporal design families and ST-MOI.
  \textit{\textbf{Right}:} Grad-CAM activation maps of three baselines
(UniFormer-S, VideoMamba-S, VideoMAE-S) and our VideoMogaNet-S
on the class \textit{Tearing something into two pieces},
suggesting sharper and more action-focused activations for our model.}
  \label{fig:arch_videomoganet_families}
\end{figure*}

\subsection{From Images to Video: Two Complementary Challenges}
\label{subsec:challenges}

MogaNet~\cite{li2022moganet} effectively captures multi-order spatial
interactions within each frame, encoding interactions of all orders
while privileging middle-order ones. However, this aggregation operates
independently on each frame $t \in \{1, \dots, T\}$, treating the
video as a collection of static images. Unlike images, videos contain
inter-frame dynamics: semantics evolve along the temporal dimension
$T$, carrying action and motion information that cannot be captured by
purely intra-frame processing. A first challenge therefore consists in
designing a temporal modeling module capable of capturing these
inter-frame dynamics at different interaction orders, at lightweight
cost so as not to negate the efficiency gains of spatial aggregation.

A second challenge, independent of the first, concerns the
spatio-temporal fusion scheme. Existing spatio-temporal fusion designs
in the literature are limited to a single interaction between the two
dimensions, typically of the form:
\begin{equation}
    \label{eq:standard_fusion}
    Z = \mathrm{spatial}(V) + \alpha \cdot \mathrm{temporal}
    \big(\mathrm{spatial}(V)\big),
\end{equation}
where $V \in \mathbb{R}^{C \times T \times H \times W}$ denotes the
input video features, $\alpha$ is a scaling factor and
$\mathrm{spatial}(\cdot)$, $\mathrm{temporal}(\cdot)$ denote the
spatial and temporal processing modules respectively. This design
allows only a single interaction between the two dimensions, leaving
unexploited some information arising from their multiplicative
interaction.

These two challenges motivate our two complementary contributions.
\textbf{MOTM} (\textit{Multi-Order Temporal Modulation}) addresses the
first challenge by proposing a multi-order temporal modeling module at
lightweight cost (${\sim}$0.282~GFLOPs), based on parallel
channel-split processing with depthwise-only convolutions and implicit
gating-based aggregation, without internal residual connection or
explicit cross-channel aggregation projections. \textbf{ST-MOI}
(\textit{Spatio-Temporal Multi-Order Interaction}) addresses the second
challenge by introducing a dual-level interaction design between the
spatial and temporal dimensions, outperforming the standard
single-interaction fusion design.

\subsection{Overall Architecture}
\label{subsec:architecture}

Taking the merits of ViT-style framework
design~\cite{yu2022metaformer}, Modern ConvNets for video action
recognition~\cite{liu2022convnext, huang2021tada} adopt the
macro-architecture of Vision Transformers while preserving the local
inductive biases of convolutions. VideoMogaNet follows this paradigm
by adopting a four-stage hierarchy, similar to~\cite{liu2022video},
where each stage $i$ comprises an \textit{embedding stem} followed by
$N_i$ \textit{VideoMoga} blocks. Unlike backbones that apply a constant doubling factor at every stage ($C_i = 2C_{i-1}$), VideoMogaNet adopts a variable
channel growth ratio across stages, reducing
the overall computational cost. Given an input
$V \in \mathbb{R}^{C \times T \times H \times W}$, where $C$, $T$,
$H$, and $W$ denote the number of channels, temporal length, height,
and width respectively, the \textit{embedding stem} produces:
\begin{equation}
    Z = \mathrm{Stem}(V),
\end{equation}
where $Z$ denotes the downsampled features, applied independently on
each frame.

Within each stage, the network modules are decoupled into four
distinct functional components. Standard Modern ConvNets apply
$\mathrm{TMixer}(\cdot)$ before $\mathrm{CMixer}(\cdot)$. VideoMogaNet reverses this order: $\mathrm{TMixer}(\cdot)$, which
captures temporal dynamics across frames, is applied \textit{after}
$\mathrm{CMixer}(\cdot)$, which performs inter-channel aggregation.
This lets temporal modulation operate on features already enriched
with spatial and inter-channel information. VideoMogaNet further
introduces a fifth component, the \textbf{STMixer}, which adaptively
fuses spatial and temporal representations through bounded
multiplicative modulation, as illustrated in
Fig.~\ref{fig:overall_videomoganet} (\textbf{left}). For an input
$V \in \mathbb{R}^{C \times T \times H \times W}$, the forward pass
of the VideoMoga block, shown in Fig.~\ref{fig:overall_videomoganet}
(\textbf{right}), is:
\begin{align}
    \label{eq:smixer}
    Y &= V + \mathrm{SMixer}\big(\mathrm{Norm}_1(V)\big), \\
    \label{eq:cmixer}
    \tilde{Y} &= Y + \mathrm{CMixer}\big(\mathrm{Norm}_2(Y)\big), \\
    \label{eq:tmixer}
    M &= \mathrm{TMixer}(\tilde{Y}), \\
    \label{eq:stmixer}
    Z &= \tilde{Y} \odot \big(1 + \alpha \cdot \tanh(M)\big),
\end{align}
where $\mathrm{Norm}(\cdot)$ denotes a normalization layer
(\textit{e.g.}, BatchNorm~\cite{ioffe2015batch}), $\odot$ denotes
element-wise multiplication, $\alpha \in \mathbb{R}$ is a learnable
scaling factor initialized to zero for progressive training stability,
and $\tanh$ bounds the modulation within $[-1, 1]$. The term
$\mathrm{STMixer} \equiv \tilde{Y} \odot (1 + \alpha \cdot \tanh(M))$
in Eq.~(\ref{eq:stmixer}) realizes our dual-level ST-MOI fusion
design, in contrast with the standard single-interaction fusion of
Eq.~(\ref{eq:standard_fusion}). In VideoMogaNet, $\mathrm{SMixer}$
instantiates the multi-order spatial gated aggregation introduced in ~\cite{li2022moganet}, $\mathrm{CMixer}$ performs
inter-channel aggregation via an inverted bottleneck
MLP~\cite{cvpr2018mobilenetv2}, and $\mathrm{TMixer}$ captures
multi-order temporal dynamics through our MOTM, detailed in
Sec.~\ref{subsec:motm}.

\subsection{Multi-Order Temporal Modulation}
\label{subsec:motm}

We introduce the MOTM module
as an instantiation of $\mathrm{TMixer}(\cdot)$, designed to capture
inter-frame temporal dynamics at different interaction orders at
lightweight computational cost. Given an input
$V \in \mathbb{R}^{BT \times C \times H \times W}$, where $B$, $T$,
$C$, $H$, and $W$ denote the batch size, temporal length, number of
channels, height, and width respectively, MOTM produces a modulation
signal $M \in \mathbb{R}^{BT \times C \times H \times W}$ that is
consumed by ST-MOI (Sec.~\ref{subsec:st_interaction}). Residuality is
deliberately handled at the ST-MOI fusion level rather than internally,
preserving the modularity of the module and avoiding redundant
residual paths. The module decomposes into three components.

\noindent \textbf{Dense Temporal Decomposition}
To process each spatial position independently along the temporal
dimension, the input $V$ is first reshaped from
$\mathbb{R}^{BT \times C \times H \times W}$ into
$\mathbb{R}^{BHW \times T \times C}$, so that each spatial position
$(h, w)$ is treated as an independent temporal sequence of length $T$.
A layer normalization followed by an adaptive temporal readjustment is
then applied:
\begin{align}
    \label{eq:td_ln}
    y &= \mathrm{LN}(v), \\
    \label{eq:td_gamma}
    z &= y + \gamma_t \odot \big(y - \bar{y}_T\big),
\end{align}
where $v \in \mathbb{R}^{BHW \times T \times C}$ denotes the reshaped
input, $\mathrm{LN}(\cdot)$ denotes Layer Normalization applied over
the channel dimension $C$, $\gamma_t \in \mathbb{R}^C$ is a learnable
scaling factor initialized to zero, and $\bar{y}_T$ denotes the
temporal mean of $y$ computed over the $T$ dimension. The term
$(y - \bar{y}_T)$ forces the network against its implicit temporal
biases by excluding trivial redundant inter-frame interactions; this
exclusion of redundant inter-frame interactions mirrors the spatial
decomposition principle introduced in~\cite{li2022moganet}, which
subtracts globally shared activations to increase feature diversity
and counter implicit interaction biases.

\begin{figure*}[!t]
\centering
\begin{subfigure}[t]{0.49\linewidth}
    \centering
    \includegraphics[width=\linewidth]{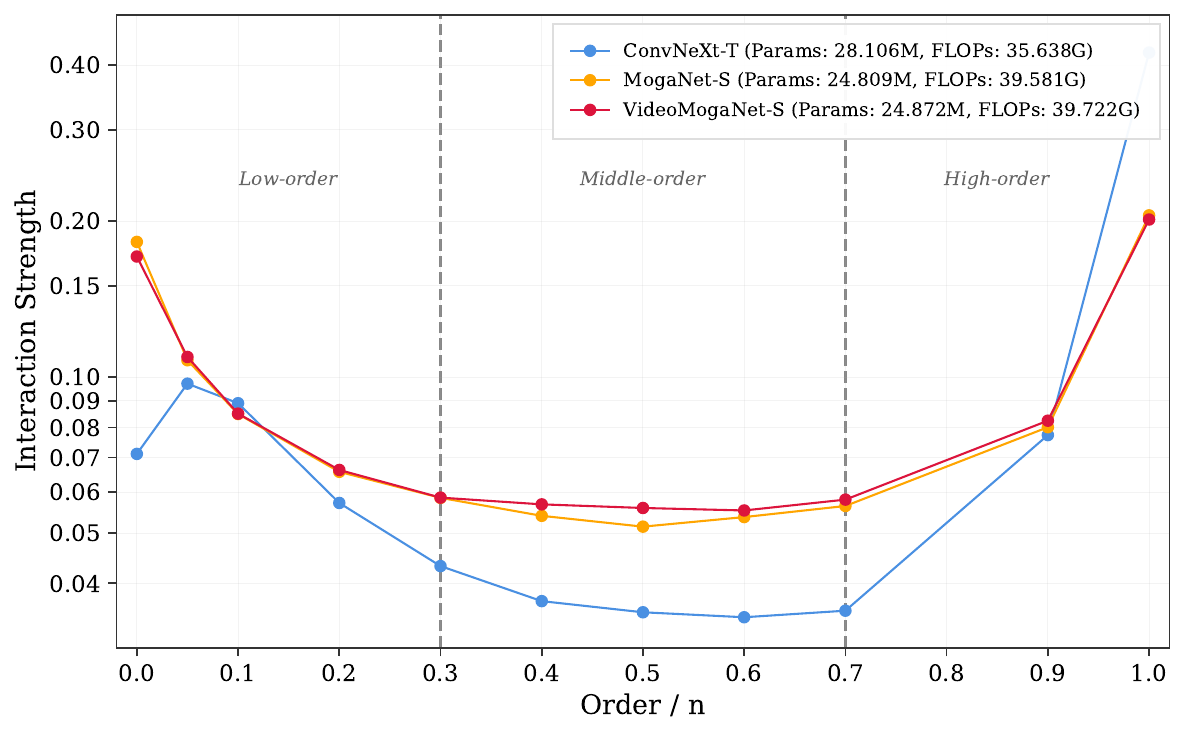}
\end{subfigure}
\hfill
\begin{subfigure}[t]{0.49\linewidth}
    \centering
    \includegraphics[width=\linewidth]{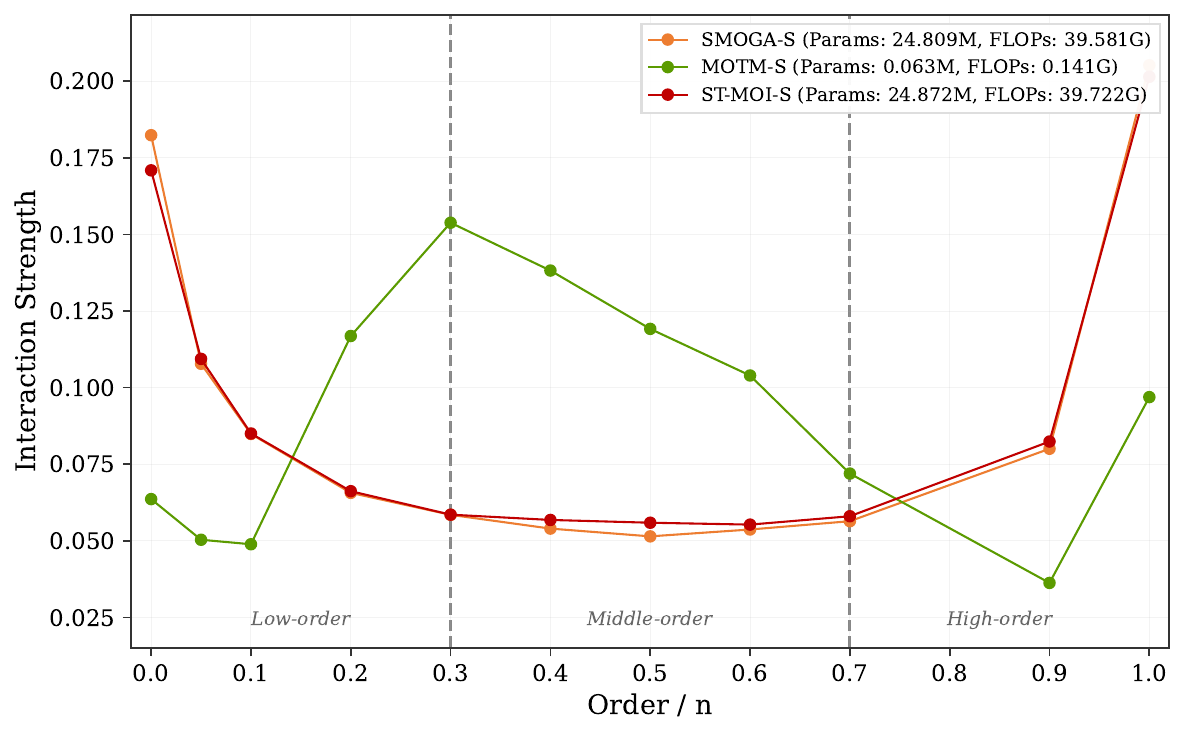}
\end{subfigure}
\caption{\textbf{Interaction strength}. \emph{\textbf{Left}:} Multi-order
interaction strength across architectures. \emph{\textbf{Right}:} Interaction
strength per component.}
\label{fig:multi_order_and_st_ablation}
\end{figure*}

\noindent \textbf{Parallel Multi-Order Temporal Extraction.}
The normalized sequence $z$ is permuted to
$z \in \mathbb{R}^{BHW \times C \times T}$ and decomposed into three
channel groups according to a $1\!:\!3\!:\!4$ ratio:
\begin{equation}
    z_0 \in \mathbb{R}^{BHW \times C_0 \times T},
    z_1 \in \mathbb{R}^{BHW \times C_1 \times T},
    z_2 \in \mathbb{R}^{BHW \times C_2 \times T}
\end{equation}
where $C_0 + C_1 + C_2 = C$. Each group is processed in parallel by a
dedicated 1D depthwise convolution (DWConv1D):
\begin{align}
    \label{eq:motm_low}
    z^{(\text{low})}  &= \mathrm{DWConv1D}_{k=5,\,d=1}(z_0), \\
    \label{eq:motm_mid}
    z^{(\text{mid})}  &= \mathrm{DWConv1D}_{k=5,\,d=2}(z_1), \\
    \label{eq:motm_high}
    z^{(\text{high})} &= \mathrm{DWConv1D}_{k=7,\,d=3}(z_2),
\end{align}
capturing respectively short transitions between adjacent frames
(low-order), actions and movements spanning multiple frames
(middle-order), and long-range temporal dependencies (high-order).
Each DWConv1D operates with $\mathrm{groups}=C_i$, ensuring a purely
depthwise $\mathcal{O}(C)$ cost. This parallel channel-split design
differs fundamentally from the sequential spatial design of
MogaNet~\cite{li2022moganet}, where a first depthwise convolution
processes all $C$ channels before factorizing its output for higher
orders, requiring cross-channel aggregation of $\mathcal{O}(C^2)$
cost. Here, each order operates on a dedicated channel partition from
the outset, eliminating expensive cross-channel projections while
preserving multi-order expressiveness.

\noindent \textbf{Adaptive Gating Aggregation.}
The multi-order representations are fused via a double gating
mechanism. Let $\mathrm{Gate}_A$ and $\mathrm{Gate}_B$ be two grouped
1D convolutions with $\mathrm{kernel}=1$ and $\mathrm{groups}=C$,
initialized with near-zero weights. The modulation signal is computed
as:
\begin{equation}
\label{eq:motm_gate}
\begin{aligned}
M &= \underbrace{\mathrm{SiLU}\!\big(\mathrm{Gate}_A(z)\big)}
        _{\mathcal{F}_{\phi'}} \odot \\
  &\quad
    \underbrace{\mathrm{SiLU}\!\big(
    \mathrm{Gate}_B(
    [z^{(\text{low})},\, z^{(\text{mid})},\,
    z^{(\text{high})}])
    \big)}
    _{\mathcal{G}_{\psi'}} .
\end{aligned}
\end{equation}
where $[\cdot,\cdot]$ denotes concatenation along the channel
dimension, $\odot$ denotes element-wise multiplication, and
$\mathrm{SiLU}$~\cite{elfwing2018sigmoid} combines gating effect with
training stability. $\mathrm{Gate}_A$ operates on the normalized
sequence $z$ as the aggregation branch $\mathcal{F}_{\phi'}$, while
$\mathrm{Gate}_B$ operates on the concatenated multi-order features as
the context branch $\mathcal{G}_{\psi'}$. The output
$M \in \mathbb{R}^{BHW \times C \times T}$ is finally reshaped back
to $\mathbb{R}^{BT \times C \times H \times W}$, fully reconstructing
the spatial structure for subsequent ST-MOI fusion.

\subsection{Spatio-Temporal Multi-Order Interaction}
\label{subsec:st_interaction}

We introduce ST-MOI as an instantiation of $\mathrm{STMixer}(\cdot)$,
fusing the spatial and temporal representations produced by
$\mathrm{SMixer}(\cdot)$ and $\mathrm{TMixer}(\cdot)$ respectively.
ST-MOI operates on $\tilde{Y}$, the post-\(\mathrm{CMixer}\) features
defined in Eq.~(\ref{eq:cmixer}), not on the raw input $V$.
The key design question is not merely whether to incorporate temporal
information, but \textit{how} spatial and temporal representations
should interact.

\vspace{+.2em}
\noindent \textbf{Dual-Level Interaction Design.}
Standard spatio-temporal fusion adopts an additive coupling:
\begin{equation}
    \label{eq:standard_st}
    Z = \tilde{Y} + \alpha \cdot \mathrm{TMixer}(\tilde{Y}),
\end{equation}
where $\alpha \in \mathbb{R}$ is a learnable scaling factor.
This allows only a single, additive interaction between spatial and
temporal dimensions, leaving their multiplicative interaction
unexploited. Fig.~\ref{fig:arch_videomoganet_families} (\textbf{left})
contrasts this standard design with our alternative.
ST-MOI instead adopts a variable-gain modulation:
\begin{equation}
    \label{eq:stmoi}
    Z = \tilde{Y} + \alpha \cdot \tilde{Y} \odot
        \tanh\!\bigl(\mathrm{TMixer}(\tilde{Y})\bigr),
\end{equation}
where $\odot$ denotes element-wise multiplication, $\alpha \in \mathbb{R}$
is a learnable scaling factor, and $\tanh(\cdot)$ bounds the modulation
within $[-1,1]$ for training stability.
The additional term $\tilde{Y} \odot \tanh\!\bigl(\mathrm{TMixer}(\tilde{Y})\bigr)$,
absent from Eq.~(\ref{eq:standard_st}), lets the temporal signal directly
modulate the spatial features in a content-dependent manner,
realizing a dual-level spatio-temporal interaction.
Eq.~(\ref{eq:stmoi}) is algebraically equivalent to Eq.~(\ref{eq:stmixer})
of the VideoMoga block, ensuring full consistency between the two formulations.
The superiority of this design over standard additive fusion is
empirically validated in Sec.~\ref{sec:experiments}.
Fig.~\ref{fig:arch_videomoganet_families} (\textbf{right}) illustrates, via Grad-CAM maps on SSv2, that VideoMogaNet produces
more accurate and temporally consistent activations on salient action
regions than UniFormer-S, VideoMamba-S, and VideoMAE-S.

\begin{table*}[t]
  \centering
  \begin{minipage}[t]{0.54\linewidth}
    \centering
    \footnotesize
    \renewcommand{\arraystretch}{.82}
    \captionof{table}{Comparison with state of the art on HMDB51 \& UCF101.
    $\dagger$: results reproduced by~\cite{li2024mamba}.}
    \label{tab:ucf_hmdb}
    \resizebox{\linewidth}{!}{
    \begin{tabular}{lccccc}
    \toprule
    Method & Pretrain & F$\times$Cl$\times$Cr & TFLOPs & HMDB51[\%] & UCF101[\%] \\
    \midrule
    VideoSwin-T$^\dagger$~\cite{liu2022video}       & IN-1K & $32\times4\times3$  & --    & 53.0 & 88.3 \\
    VideoSwin-S$^\dagger$~\cite{liu2022video}       & IN-1K & $32\times4\times3$  & --    & 58.1 & 88.7 \\
    Mamba-ND~\cite{li2024mamba}                     & IN-1K & $32\times4\times3$  & --    & 60.9 & 89.6 \\
    VideoMamba-M~\cite{li2024videomamba}            & IN-1K & $16\times4\times3$  & 2.400 & 60.8 & 88.2 \\
    VideoMambaPro-M~\cite{lu2025snakes}             & IN-1K & $16\times4\times3$  & 2.200 & 63.2 & 91.6 \\
    VideoMamba~\cite{park2024videomamba}            & K400  & $16\times5\times3$  & 0.510 & 68.6 & --   \\
    VideoMAE (ViT-B)~\cite{tong2022videomae}        & K400  & $16\times10\times3$ & 5.400 & 73.3 & 96.1 \\
    \midrule
    \multicolumn{6}{c}{Modern ConvNet} \\
    \midrule
    S4ND$^\dagger$~\cite{nguyen2022s4nd}            & IN-1K & $32\times4\times3$  & --    & 56.6 & 69.3 \\
    \cellcolor{gray!20}VideoMogaNet-B (ours)        & \cellcolor{gray!20}K400 & \cellcolor{gray!20}$8\times4\times3$ & \cellcolor{gray!20}0.953 & \cellcolor{gray!20}\textbf{74.7} & \cellcolor{gray!20}\textbf{97.1} \\
    \bottomrule
    \end{tabular}}

    \vspace{.4em}

    \renewcommand{\arraystretch}{.82}
    \captionof{table}{Comparison with state of the art on Diving48.
    *: Reproduced by~\cite{kim2024learning}.
    $\dagger$: Trained with additional bbox annotations.}
    \label{tab:diving48}
    \resizebox{\linewidth}{!}{
    \begin{tabular}{lcc}
    \toprule
    Method & Frame$\times$Crop$\times$Clip & Top-1[\%] \\
    \midrule
    SlowFast-R101~\cite{feichtenhofer2019slowfast}   & $(64{+}16)\times3\times1$ & 77.6 \\
    AIA(TSM)~\cite{hao2022attention}                 & $8\times1\times1$         & 79.4 \\
    SViT-DD~\cite{ben2022bringing}                   & --                        & 79.8 \\
    TimeSformer-L~\cite{bertasius2021space}           & $96\times3\times1$        & 81.0 \\
    TQN~\cite{zhang2021temporal}                     & all frames                & 81.8 \\
    RSANet-R50~\cite{rsa}                            & $16\times1\times2$        & 84.2 \\
    VIMPAC~\cite{tan2021vimpac}                      & $-\times3\times10$        & 85.5 \\
    UniFormer-B*~\cite{uniformer}                    & $32\times3\times1$        & 87.4 \\
    StructViT-B-4-2~\cite{kim2024learning}           & $32\times3\times1$        & 87.8 \\
    GC-TDN~\cite{hao2022group}                       & $16\times1\times1$        & 87.6 \\
    ORViT TimeSformer$^\dagger$~\cite{herzig2022object} & $32\times3\times1$     & 88.0 \\
    TFCNet~\cite{zhang2022tfcnet}                    & $32\times3\times1$        & 88.3 \\
    StructViT-B-4-1~\cite{kim2024learning}           & $32\times3\times1$        & 88.3 \\
    PosMLP-Video-L~\cite{hao2024posmlp}              & $16\times1\times1$        & 88.9 \\
    \midrule
    \multicolumn{3}{c}{Modern ConvNet} \\
    \midrule
    \cellcolor{gray!20}VideoMogaNet-B (ours) & \cellcolor{gray!20}$16\times1\times1$ & \cellcolor{gray!20}\textbf{90.30} \\
    \bottomrule
    \end{tabular}}
  \end{minipage}%
  \hfill
  \begin{minipage}[t]{0.43\linewidth}
    \centering
    \footnotesize
    \renewcommand{\arraystretch}{.82}
    \captionof{table}{Comparison with state of the art on K400.}
    \label{tab:k400}
    \resizebox{\linewidth}{!}{
    \begin{tabular}{lcccc}
    \toprule
    Method & Pretrain & Frames & TFLOPs & Top-1[\%] \\
    \midrule
    CoViAR~\cite{wu2018compressed}              & IN-1K & 8     & --     & 69.1 \\
    R(2+1)D~\cite{tran2018closer}               & --    & 32    & 0.750  & 72.0 \\
    I3D~\cite{carreira2017quo}                  & IN-1K & 32    & 3.240  & 72.1 \\
    TEAM-Net~\cite{wang2021team}                & IN-1K & 8     & --     & 72.2 \\
    MViT-S~\cite{fan2021multiscale}             & --    & 16    & 0.165  & 76.0 \\
    VideoMamba~\cite{park2024videomamba}        & IN-1K & 16    & 0.510  & 76.1 \\
    VideoMamba~\cite{park2024videomamba}        & IN-1K & 32    & 1.020  & 77.7 \\
    NL I3D~\cite{wang2018non}                   & IN-1K & 32    & 10.770 & 77.7 \\
    MViT-B~\cite{fan2021multiscale}             & --    & 16    & 0.355  & 78.4 \\
    VideoSwin-T~\cite{liu2022video}             & IN-1K & 32    & 1.056  & 78.8 \\
    VideoMAE (ViT-S)~\cite{tong2022videomae}    & --    & 16    & 0.855  & 79.0 \\
    SlowFast~\cite{feichtenhofer2019slowfast}   & --    & 16+64 & 7.020  & 79.8 \\
    VideoMAE (ViT-L)~\cite{tong2022videomae}    & --    & 16    & 8.955  & 85.2 \\
    VideoMAE (ViT-H)~\cite{tong2022videomae}    & --    & 16    & 17.880 & 86.6 \\
    VideoMAE-V2 (ViT-H)~\cite{wang2023videomae} & Multi & 16   & 17.880 & 88.6 \\
    UniFormerV2-L~\cite{li2023uniformerv2}      & CLIP  & 32    & 16.000 & 89.3 \\
    InternVideo-T~\cite{wang2022internvideo}    & Multi & --    & --     & 91.1 \\
    \midrule
    \multicolumn{5}{c}{Modern ConvNet} \\
    \midrule
    Das~\textit{et al.}-B~\cite{das2026learning}   & IN-1K & 8  & --    & 74.2 \\
    Das~\textit{et al.}-L~\cite{das2026learning}   & IN-1K & 8  & --    & 75.9 \\
    TAdaConvNeXt-T~\cite{huang2021tada}            & IN-1K & 32 & 1.128 & 79.1 \\
    \cellcolor{gray!20}VideoMogaNet-B (ours) & \cellcolor{gray!20}IN-1K & \cellcolor{gray!20}8 & \cellcolor{gray!20}0.953 & \cellcolor{gray!20}80.6 \\
    \cellcolor{gray!20}VideoMogaNet-L (ours) & \cellcolor{gray!20}IN-1K & \cellcolor{gray!20}8 & \cellcolor{gray!20}1.527 & \cellcolor{gray!20}\textbf{81.3} \\
    \bottomrule
    \end{tabular}}

    \captionof{figure}{Per-class accuracy gain from MOTM on K400. Top-5 classes
    benefiting most from MOTM (positive) and least (negative).}
    \label{fig:motm_gain_k400}
    \includegraphics[width=\linewidth, height=4.5cm, keepaspectratio]{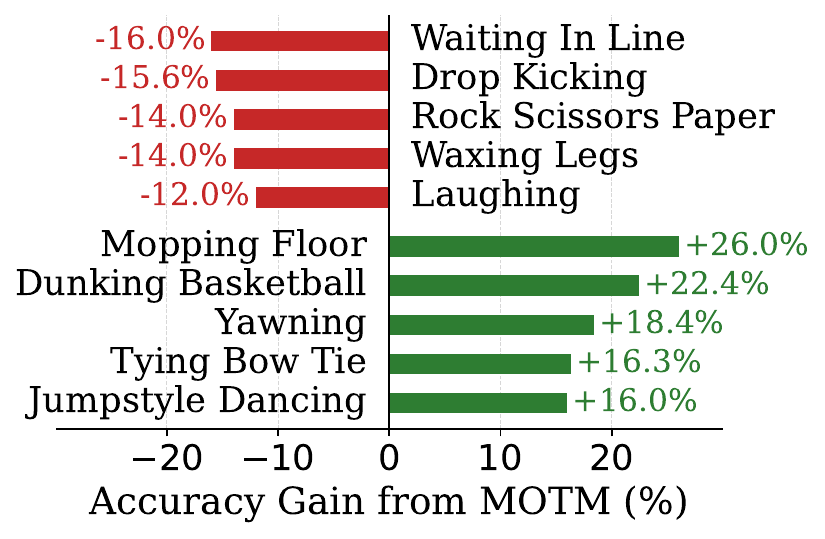}
  \end{minipage}
\end{table*}


\section{Multi-Order Interaction Strength}
\label{subsec:interaction}
\noindent \textbf{Experimental Protocol.}
We train three architectures on the Kinetics-400 training set and
evaluate interaction strength on its validation set:
(i) ConvNeXt-T with temporal frame averaging,
(ii) MogaNet-S with temporal averaging~\cite{li2022moganet}, and
(iii) VideoMogaNet-S.
For each video, we define a spatio-temporal grid of
$T \times H \times W = 8 \times 14 \times 14$ patches.
We generate 200 patch pairs within a 3D local neighborhood
and sample 20 random contexts per interaction order $m$, following
the approximation protocol of~\cite{deng2021discovering,
zhang2020interpreting}. The relative interaction strength $J^{(m)}$
is computed for each architecture and averaged across videos and
patch pairs.

\vspace{.4em}
\noindent \textbf{Analysis of Interaction Strength.}
Fig.~\ref{fig:multi_order_and_st_ablation} (\textbf{left}) compares
the relative interaction strength
$J^{(m)}$~\cite{deng2021discovering,zhang2020interpreting}, which
measures the normalized contribution of interactions of order $m$
($m{=}0$: low-order; $m{=}0.5$: middle-order; $m{=}1$: high-order),
across three architectures.
While all three architectures share a broadly similar profile,
favoring extreme orders with a relative trough at middle orders,
consistent differences emerge.
ConvNeXt+Avg exhibits the weakest middle-order strength
($J^{(0.5)}{=}0.035$).
MogaNet+Avg progressively shifts interaction mass toward middle orders
($J^{(0.5)}{=}0.051$, $+46\%$ vs ConvNeXt+Avg), yet remains
constrained by temporal averaging.
VideoMogaNet further strengthens this trend
($J^{(0.5)}{=}0.056$, $+59\%$ vs ConvNeXt+Avg, $+9\%$ vs
MogaNet+Avg), suggesting that ST-MOI coordination progressively
amplifies middle-order interactions across both spatial and
temporal dimensions.

\vspace{.2em}
\noindent \textbf{Per-Component Interaction Profile.}
Fig.~\ref{fig:multi_order_and_st_ablation} (\textbf{right})
decomposes ST-MOI into its individual components.
SMixer (SMOGA) encodes interactions of all orders within each frame
while moderately favoring middle-order ones ($J^{(0.5)}{=}0.051$),
consistent with~\cite{li2022moganet}.
MOTM concentrates its interaction mass in the middle-order range
($m{=}0.3$--$0.5$), peaking at $m{=}0.3$ ($J^{(0.3)}{=}0.154$) and
remaining elevated at $m{=}0.5$ ($J^{(0.5)}{=}0.119$), reflecting
that multi-order temporal modulation is more selectively tuned to
medium temporal spans than its spatial counterpart.
ST-MOI coordinates these two components through bounded multiplicative
modulation, yielding a distribution that largely preserves the spatial
profile of SMixer (SMOGA) while integrating the temporal dynamics of
MOTM, with a modest but consistent gain in middle-order strength
($J^{(0.5)}{=}0.056$ vs $0.051$).
These results are consistent with the complementarity of the three
components, though the relatively similar overall profiles across
architectures suggest that the gains are gradual rather than
qualitatively distinct.

\section{Results and Discussion}
\label{sec:experiments}

\subsection{Experimental Setup}
\label{subsec:setup}

\noindent \textbf{Datasets.}
We evaluate VideoMogaNet on four popular video benchmarks.
\textbf{UCF-101}~\cite{soomro2012ucf101} contains 13,320 trimmed videos
from 101 action categories, while \textbf{HMDB-51}~\cite{kuehne2011hmdb}
contains 6,766 videos from 51 categories; both datasets provide three
training-testing splits, and we report the average top-1 accuracy
across the 3 splits. \textbf{Kinetics-400 (K400)}~\cite{kay2017k400}
comprises approximately 240k training videos and 20k validation videos
across 400 action classes. \textbf{Diving-48 (D-48)}~\cite{yingwei2018d48}
is a challenging dataset of diving actions consisting of
${\sim}$15,000 training and ${\sim}$2,000 testing samples. Actions are
differentiated only by the subtle movement of the diver across frames,
with the background remaining largely constant, requiring robust
temporal modeling for accurate recognition. As shown
by~\cite{wang2021bevt}, disrupting temporal information on this
dataset, either through random shuffling or single-frame evaluation,
causes accuracy drops of up to ${\sim}33.6\%$ and ${\sim}70.2\%$
respectively, underlining the importance of temporal modeling on this
benchmark.

\vspace{.2em}
\noindent \textbf{Implementation Details.}
All models are initialized with ImageNet-1K pre-trained
weights~\cite{li2022moganet}. We use the AdamW optimizer with a cosine
annealing learning rate schedule and linear warmup. An effective batch
size of 256 is maintained across all experiments via gradient
accumulation. The learning rate is set to $8 \times 10^{-4}$ for
UCF-101, HMDB-51, K400, and D-48. Training epochs vary by dataset and
model variant: on UCF-101, HMDB-51, and D-48, Base trains for 120
epochs; on K400, Base/Large train for 30 epochs, Small for 50, and
Tiny for 90. Training is conducted on 4$\times$ NVIDIA A100-SXM4-40GB
GPUs with mixed-precision training (AMP O1, FP16). All evaluations use the spatial resolution $224^2$, with 8-frame input clips for UCF-101, HMDB-51, and K400, and 16-frame input clips
for D-48. Final predictions are obtained by averaging softmax
probabilities across all views. FLOPs depend on dataset because
evaluation protocols differ: UCF-101, HMDB-51, and K400 require 12
forward passes ($4{\times}3$ views) while D-48 requires a single
forward pass ($1{\times}1$ view).

\subsection{Qualitative Analysis of MOTM}
\label{subsec:qualitative}

Comparing VideoMogaNet vs.\ VideoMogaNet$\setminus$MOTM on K400
reveals that MOTM benefits actions with strong temporal dynamics
(\textit{mopping floor} $+26\%$, \textit{dunking basketball} $+22\%$),
confirming its motivation. Occasional drops
(\textit{waiting in line} $-16\%$) reflect either classes where spatial
context dominates, or noise induced by class imbalance in K400
(Fig.~\ref{fig:motm_gain_k400}).

\subsection{Quantitative Results}
\label{subsec:comparison}

\noindent \textbf{UCF-101.}
VideoMogaNet achieves the best accuracy among the compared methods on
UCF-101 (Tab.~\ref{tab:ucf_hmdb}), reaching $97.1\%$ top-1. It
outperforms VideoMAE~\cite{tong2022videomae} ($96.1\%$), the strongest of
the other methods, by $1.0\%$ at a lower total cost ($0.953$T with $12$
views versus $5.4$T with $30$ views), and
VideoMambaPro-M~\cite{lu2025snakes} ($91.6\%$) by $5.5\%$.

\noindent \textbf{HMDB-51.}
On HMDB-51 (Tab.~\ref{tab:ucf_hmdb}), VideoMogaNet reaches
$74.7 \pm 0.45\%$ averaged over three splits, the best accuracy among the
compared methods. It outperforms
VideoMAE~\cite{tong2022videomae} ($73.3\%$, $+1.4\%$) at a $5.7{\times}$
lower total cost, and VideoMamba~\cite{park2024videomamba} ($68.6\%$,
$+6.1\%$).

\noindent \textbf{Kinetics-400.}
On K400 (Tab.~\ref{tab:k400}), VideoMogaNet delivers competitive
performance at comparable cost rather than the largest possible
margin. VideoMogaNet-L reaches $81.3\%$, surpassing our baseline,
Das~\textit{et al.}~\cite{das2026learning}-L ($75.9\%$), by $5.4\%$,
and outperforming SlowFast~\cite{feichtenhofer2019slowfast}
($79.8\%$), TAdaConvNeXt-T~\cite{huang2021tada} ($79.1\%$), and
VideoSwin-T~\cite{liu2022video} ($78.8\%$). Against
VideoMamba~\cite{park2024videomamba} ($77.7\%$ at $1.020$T), the
closest method in terms of cost, VideoMogaNet-L holds a $3.6\%$ lead
at a comparable $1.527$T.

\vspace{.4em}
\noindent \textbf{Diving-48.}
Tab.~\ref{tab:diving48} shows that VideoMogaNet establishes the best
result among all compared methods on Diving-48 at $90.30\%$ top-1
accuracy, surpassing PosMLP-Video-L~\cite{hao2024posmlp} ($88.9\%$),
the previous state of the art on this dataset, by $1.4\%$. Since
actions on Diving-48 are distinguished only by the subtle, frame-to-frame
motion of the diver against an otherwise static background, this
result highlights VideoMogaNet's ability to model fine-grained
temporal dynamics, consistent with the role of MOTM in capturing
inter-frame motion at lightweight cost.

\subsection{Complexity Analysis}
\label{subsec:complexity}

\begin{table}[t]
\centering
\footnotesize
\setlength{\tabcolsep}{3pt}
\renewcommand{\arraystretch}{0.88}
\caption{Asymptotic complexity. $n_t$: temporal length;
$n_h \times n_w$: spatial resolution; $d$: channel dimension.}
\label{tab:complexity}
\begin{tabular}{lcc}
    \toprule
    Architecture & Complexity & Scaling in $n_t$ \\
    \midrule
    ViViT~\cite{arnab2021vivit}          & $\mathcal{O}\big((n_h n_w)^2 d + n_t^2 d\big)$ & Quadratic \\
    VideoMamba~\cite{park2024videomamba} & $\mathcal{O}(n_h n_w n_t d)$                    & Linear \\
    \midrule
    \textbf{VideoMogaNet} (total)        & $\mathcal{O}(n_t n_h n_w d^2)$                  & Linear \\
    \textbf{MOTM} (ours)                 & $\mathcal{O}(n_t n_h n_w d)$                    & Linear, no $d^2$ \\
    \bottomrule
    \end{tabular}
\end{table}

\noindent \textbf{Complexity of MOTM.}
For an input of size $B \times T \times C \times H \times W$, MOTM treats
each spatial position as an independent temporal sequence through a
reshape that costs no FLOPs. The three $\mathrm{DWConv1D}$ operate on
disjoint channel groups and cost $11.5\,BHWCT$ FLOPs, while the gating
(two grouped $\mathrm{Conv1d}$ with $\mathrm{groups}{=}C$) adds
$4\,BHWCT$:
\begin{equation}
    \label{eq:cplx_motm}
    F_{\mathrm{MOTM}} = 15.5 \cdot B \cdot H \cdot W \cdot C \cdot T
    = \mathcal{O}(BTHWC).
\end{equation}
No cross-channel projection is involved, hence no $C^2$ term. ST-MOI only
adds four element-wise operations, \textit{i.e.} $4\,BTHWC$ FLOPs, for a
single parameter ($\alpha$).

\noindent \textbf{Complexity of VideoMogaNet.}
The cost of a VideoMoga block is
$F_{\mathrm{Sp}} + F_{\mathrm{MOTM}} + F_{\mathrm{ST\text{-}MOI}}$,
where the spatial term inherited from the backbone is
$F_{\mathrm{Sp}} = T\,HWC\,(22C + 189.75) \approx 22\,THWC^2$. The cost of
the full network is therefore dominated by this term:
\begin{equation}
    \label{eq:cplx_net}
    \mathrm{FLOPs} \approx 22 \cdot T \cdot \sum_{i=1}^{4} N_i\, H_i\, W_i\, C_i^2 .
\end{equation}
The relative temporal overhead
$\varepsilon = 15.5\,B\,/\,(22C + 189.75) = \mathcal{O}(1/C)$ decreases with
model width and becomes negligible for larger models.

\noindent \textbf{Comparison.}
Tab.~\ref{tab:complexity} compares the asymptotic complexity of
VideoMogaNet with that of the other architectures. ViViT~\cite{arnab2021vivit}
is quadratic in $n_t$ and in spatial resolution, whereas
VideoMamba~\cite{park2024videomamba} is linear thanks to state space
models. VideoMogaNet is linear in $n_t$, $n_h$ and $n_w$: it inherits a
$d^2$ term from its spatial backbone, required for multi-order
cross-channel interaction, while MOTM and ST-MOI are strictly linear.

\begin{figure*}[t]
\centering
\begin{minipage}[t]{0.37\linewidth}
    \centering
    \includegraphics[width=\linewidth]{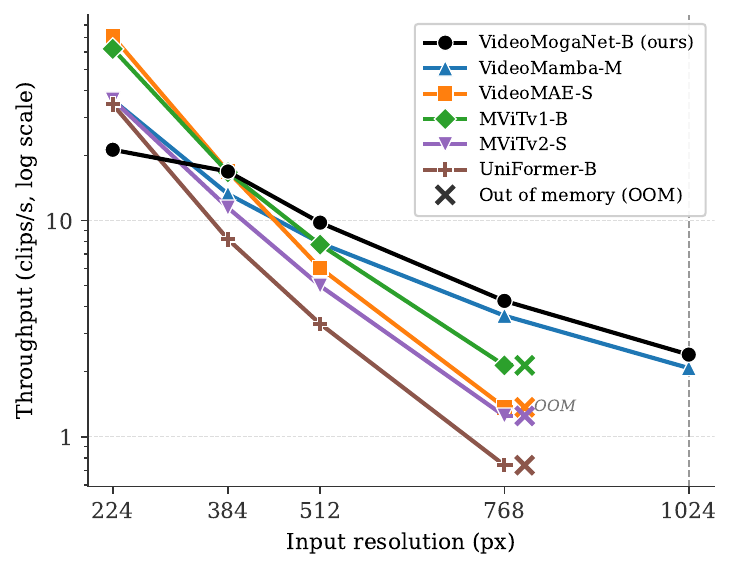}
    \subcaption{}
\end{minipage}
\hfill
\begin{minipage}[t]{0.30\linewidth}
    \centering
    \includegraphics[width=\linewidth]{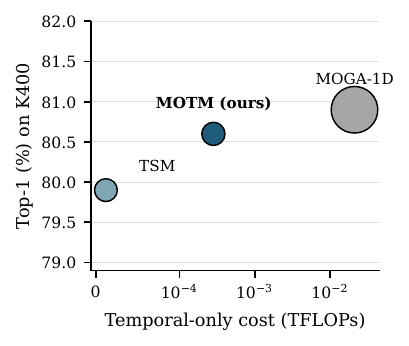}
    \subcaption{}
\end{minipage}
\hfill
\begin{minipage}[t]{0.30\linewidth}
    \centering
    \includegraphics[width=\linewidth]{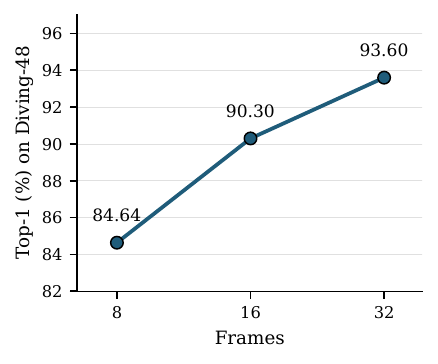}
    \subcaption{}
\end{minipage}
\caption{\textbf{Ablation analyses.} \textbf{(a)} Inference throughput
versus input resolution across six architectures. \textbf{(b)} Temporal
modules on K400. \textbf{(c)} Temporal scalability of VideoMogaNet-B on Diving-48 at 8, 16 and 32 frames.}
\label{fig:ablation_analyses}
\end{figure*}

\begin{table}[t]
\centering
\footnotesize
\setlength{\tabcolsep}{3pt}
\renewcommand{\arraystretch}{0.88}
\caption{ST-MOI and mixer ordering ablations on Diving-48.}
\label{tab:fusion_mixer}
\begin{tabular}{lc}
    \toprule
    Design & Top-1 \\
    \midrule
    \multicolumn{2}{l}{\textit{Spatio-temporal fusion}} \\
    Standard (add.)        & 89.5 \\
    \rowcolor{gray!20}
    ST-MOI (ours)          & \textbf{90.3} \\
    \midrule
    \multicolumn{2}{l}{\textit{Mixer ordering}} \\
    $\mathrm{S}{\to}\mathrm{T}{\to}\mathrm{C}$        & 89.3 \\
    $\mathrm{S}{\parallel}\mathrm{T}{\to}\mathrm{C}$  & 89.7 \\
    $\mathrm{T}{\to}\mathrm{S}{\to}\mathrm{C}$        & 89.1 \\
    \rowcolor{gray!20}
    $\mathrm{S}{\to}\mathrm{C}{\to}\mathrm{T}$ (ours) & \textbf{90.3} \\
    \bottomrule
    \end{tabular}
\end{table}

\begin{table}[t]
\centering
\footnotesize
\setlength{\tabcolsep}{3pt}
\renewcommand{\arraystretch}{0.88}
\caption{MOTM ablation on Diving-48.}
\label{tab:ablation_motm}
\begin{tabular}{lc}
    \toprule
    Configuration & Top-1 \\
    \midrule
    $\mathrm{DW1d}_{5,d=1}$ (single)                                                    & 88.9 \\
    $\mathrm{DW1d}_{5,d=1}$+$\mathrm{DW1d}_{7,d=3}$ (serial)                           & 89.3 \\
    $\mathrm{DW1d}_{5,d=1}$+$\mathrm{DW1d}_{5,d=2}$+$\mathrm{DW1d}_{7,d=3}$ (serial)  & 89.7 \\
    \midrule
    Parallel, $C_l{:}C_m{:}C_h = 0{:}1{:}1$                                            & 89.6 \\
    Parallel, $C_l{:}C_m{:}C_h = 1{:}6{:}9$                                            & 89.9 \\
    \rowcolor{gray!20}
    Parallel, $C_l{:}C_m{:}C_h = 1{:}3{:}4$ (MOTM)                                     & \textbf{90.3} \\
    \bottomrule
    \end{tabular}
\end{table}

\subsection{Ablation Study}
\label{subsec:ablation}

\noindent \textbf{Inference Throughput.}
Fig.~\ref{fig:ablation_analyses} (a) reports inference throughput across
six architectures as input resolution increases from $224^2$ to
$1024^2$. VideoMogaNet-B's throughput degrades far more gracefully with
resolution than all five baselines, dropping only $5.0{\times}$ from
$224^2$ to $768^2$, versus $9.9$--$52{\times}$ for the others. This
allows VideoMogaNet-B to overtake every baseline beyond $512^2$ and
remain, together with the SSM-based VideoMamba, one of only two
architectures that do not run out of memory at $1024^2$.

\noindent \textbf{Temporal Modeling Comparison.}
Fig.~\ref{fig:ablation_analyses} (b) compares MOTM against MOGA-1D and
TSM on K400. MOTM matches TSM ($80.6\%$ vs $79.9\%$) and stays within
$0.3\%$ of MOGA-1D ($80.9\%$) at a ${\sim}70{\times}$ lower temporal
cost.

\noindent \textbf{Temporal Scalability.}
VideoMogaNet-B reaches $84.64\%$, $90.30\%$ and $93.60\%$ top-1 on Diving-48 with 8, 16 and 32 frames ($+5.66\%$, then $+3.30\%$), confirming that MOTM effectively exploits longer sequences.

\noindent \textbf{Spatio-Temporal Fusion Design.}
Tab.~\ref{tab:fusion_mixer} compares our dual-level ST-MOI fusion
against the standard additive design at equal cost. ST-MOI improves
Top-1 by $+0.8\%$ on Diving-48 ($89.5\% \rightarrow 90.3\%$), showing the
benefit of the multiplicative interaction when fine-grained temporal
dynamics are the main discriminative signal.

\noindent \textbf{Mixer Ordering.}
Applying $\mathrm{TMixer}$ after $\mathrm{CMixer}$, our design choice,
gives the best result among the four arrangements tested ($90.3\%$),
ahead of $\mathrm{S}{\parallel}\mathrm{T}{\to}\mathrm{C}$ ($89.7\%$),
$\mathrm{S}{\to}\mathrm{T}{\to}\mathrm{C}$ ($89.3\%$) and
$\mathrm{T}{\to}\mathrm{S}{\to}\mathrm{C}$ ($89.1\%$).

\noindent\textbf{MOTM Design.}
Tab.~\ref{tab:ablation_motm} validates the design choices of MOTM. The
parallel design, where each channel group captures a distinct temporal
order, outperforms its three-branch serial counterpart ($90.3\%$ vs
$89.7\%$) and a single temporal convolution ($88.9\%$). The $1{:}3{:}4$
split gives the best result among the splits tested.

\section{Conclusion}
We have presented VideoMogaNet, an efficient video recognition
architecture built on two complementary contributions: MOTM, a
multi-order temporal modeling module at lightweight cost
(${\sim}0.282$ GFLOPs), and ST-MOI, a dual-level spatio-temporal fusion
design outperforming standard additive fusion. On Diving-48, a
benchmark specifically dedicated to fine-grained temporal modeling,
VideoMogaNet reaches $90.3\%$ top-1 accuracy, surpassing
UniFormer-B~\cite{uniformer} by $2.9\%$ and
PosMLP-Video~\cite{hao2024posmlp}, the previous state of the art on
this dataset, by $1.4\%$. VideoMogaNet also reaches the best
accuracy among compared methods on HMDB-51 and UCF-101, while degrading far more gracefully in
throughput than every compared architecture, with only a $5.0{\times}$
drop from $224^2$ to $768^2$ versus $9.9$--$52{\times}$ for the
baselines. On K400, VideoMogaNet delivers competitive performance at
comparable cost. The game-theoretic analysis further confirms that
VideoMogaNet privileges middle-order interactions across both spatial
and temporal dimensions, consistent with these results.

\vspace{0.6em}
\noindent \textbf{Supplementary materials.} Provides theoretical
details of the game-theoretic framework, implementation details, and
additional hyperparameter ablations.

\section*{Acknowledgments}
This work received partial support from Distech Controls Inc., the Natural Sciences and Engineering Research Council of Canada, the Digital Research Alliance of Canada, and MITACS.


{\small
\bibliographystyle{ieee}
\bibliography{egbib}
}


\setcounter{page}{1}

\twocolumn[
\begin{center}
{\Large \bfseries Spatio-Temporal Multi-Order Interaction for Video Recognition}

\vspace{0.5em}

{\large Supplementary Material}

\vspace{1em}
\end{center}
]


\section{Architectural Details}
\label{sec:supp:arch}

\noindent This supplementary material provides additional details of VideoMogaNet as follows:

\begin{itemize}
    \item Architectural Details (\autoref{sec:supp:arch})
    \item Implementation Details (\autoref{sec:supp:impl})
    \item Multi-Order Interaction Analysis (\autoref{sec:supp:empirical})
    \item Computational Complexity Analysis (\autoref{sec:supp:complexity})
    \item Extended Experiments (\autoref{sec:supp:extended})
    \item Additional Information (\autoref{sec:supp:additional})
\end{itemize}

\subsection{Modern ConvNet Architecture for Video}
\label{sec:supp:modern_convnet}

\noindent \textbf{General Modern ConvNet architecture.} Taking the merits of ViT-style framework design~\cite{yu2022metaformer}, modern ConvNets for video action recognition~\cite{liu2022convnext, Liu2022SLak, nips2022hornet, nips2022focalnet, huang2021tada, nguyen2022s4nd} adopt the macro-architecture of Vision Transformers while preserving the local inductive biases of convolutions. Architectures such as TAdaConvNeXt~\cite{huang2021tada} notably employ large kernel depth-wise convolutions to enhance global spatio-temporal perception while maintaining computational efficiency. Such architectures primarily comprise four functional components: (\romannumeral1) embedding stem, (\romannumeral2) spatial mixing block ($\mathrm{SMixer}$), (\romannumeral3) temporal mixing block ($\mathrm{TMixer}$), and (\romannumeral4) channel mixing block ($\mathrm{CMixer}$). The embedding stem downsamples the input to reduce redundancies and computational overhead. Given an input video feature $V \in \mathbb{R}^{C \times T \times H \times W}$, where $C$, $T$, $H$, and $W$ denote the number of channels, temporal length, height, and width respectively, the stem produces:
\begin{equation}
Z = \mathrm{Stem}(V),
\end{equation}
where $Z$ denotes the downsampled features, applied independently on each frame. The feature then flows through a stack of residual blocks. In each stage, the network modules are decoupled into three separate functional components, $\mathrm{SMixer}(\cdot)$, $\mathrm{TMixer}(\cdot)$, and $\mathrm{CMixer}(\cdot)$, for spatial-wise, temporal-wise, and channel-wise information propagation:
\begin{align}
Y &= V + \mathrm{SMixer}\big(\mathrm{Norm}(V)\big), \\
\tilde{Y} &= Y + \mathrm{TMixer}\big(\mathrm{Norm}(Y)\big), \\
Z &= \tilde{Y} + \mathrm{CMixer}\big(\mathrm{Norm}(\tilde{Y})\big),
\end{align}
where $\mathrm{Norm}(\cdot)$ denotes a normalization layer, \textit{e.g.,} BatchNorm~\cite{ioffe2015batch}. $\mathrm{SMixer}(\cdot)$ and $\mathrm{TMixer}(\cdot)$ can be instantiated by various spatial and temporal operations (\textit{e.g.,} self-attention, convolution), while $\mathrm{CMixer}(\cdot)$ is typically implemented as a channel MLP with inverted bottleneck~\cite{cvpr2018mobilenetv2} and expansion ratio $r$.

\noindent \textbf{VideoMogaNet architecture.}
VideoMogaNet distinguishes itself from this general design in two
respects. First, it applies $\mathrm{TMixer}$ \textit{after}
$\mathrm{CMixer}$ rather than before, allowing temporal modulation to
operate on features already enriched with spatial and inter-channel
information. Tab.~\ref{tab:fusion_mixer} in the main paper confirms that this
ordering outperforms parallel and alternative sequential arrangements
on Diving-48.
Second, it introduces a fifth component, the \textbf{STMixer}
(Spatio-Temporal Mixer), which introduces a dual-level interaction
between spatial and temporal dimensions, whereas existing fusion
designs are limited to a single interaction. Concretely, $\mathrm{TMixer}$
in VideoMogaNet does not produce a residual output; instead, it
produces a modulation signal $M$ that is consumed by $\mathrm{STMixer}$,
whose learnable scale $\alpha \in \mathbb{R}$, initialized to zero,
ensures that training begins from a spatial-only baseline and
progressively activates the temporal branch. For an input
$V \in \mathbb{R}^{C \times T \times H \times W}$, the forward pass of
the VideoMoga block is:
\begin{align}
Y &= V + \mathrm{SMixer}\big(\mathrm{Norm}_1(V)\big), \\
\tilde{Y} &= Y + \mathrm{CMixer}\big(\mathrm{Norm}_2(Y)\big), \\
M &= \mathrm{TMixer}(\tilde{Y}), \\
Z &= \underbrace{\tilde{Y} \odot \big(1 + \alpha \cdot \tanh(M)\big)}_{\mathrm{STMixer}},
\end{align}
where $\odot$ denotes element-wise multiplication and $\tanh$ bounds
the modulation within $[-1, 1]$ to prevent gradient explosion. In
VideoMogaNet, $\mathrm{SMixer}$ instantiates the multi-order spatial
gated aggregation introduced in~\cite{li2022moganet}, $\mathrm{CMixer}$
performs inter-channel aggregation via an inverted bottleneck
MLP~\cite{cvpr2018mobilenetv2}, and $\mathrm{TMixer}$ captures
multi-order temporal dynamics through our MOTM module
(Sec.~\ref{subsec:motm} of the main paper).

\noindent \textbf{Context aggregation.}
We abstract \textit{context aggregation} in modern video ConvNets as a series of operations that \textit{adaptively} aggregate contextual information while suppressing trivial redundancies, both in spatial and temporal mixing blocks, between two embedded features. Context aggregation in spatial and temporal mixing blocks can be abstracted as adaptive fusion between aggregation and context branches, applied along complementary dimensions:
\begin{equation}
\begin{aligned}
O_{2D} &= \mathcal{S}_{2D}\!\big(\mathcal{F}_{\phi}(V),\;\mathcal{G}_{\psi}(V)\big),\\
O_{1D} &= \mathcal{S}_{1D}\!\big(\mathcal{F}_{\phi'}(V),\;\mathcal{G}_{\psi'}(V)\big),
\end{aligned}
\end{equation}
where $\mathcal{F}_{\phi}(\cdot)$ and $\mathcal{F}_{\phi'}(\cdot)$ (aggregation branches) estimate the importance of each spatial $(H \times W)$ or temporal $T$ position, $\mathcal{G}_{\psi}(\cdot)$ and $\mathcal{G}_{\psi'}(\cdot)$ (context branches) produce contextualized features, and $\mathcal{S}(\cdot,\cdot)$ denotes the fusion operator, concretely implemented as element-wise multiplication between the two branches after SiLU gating. Context aggregation models the importance of each position on $V$ by the aggregation branch and reweights the embedded feature from the context branch.

\noindent \textbf{VideoMogaNet context aggregation.}
VideoMogaNet instantiates both spatial and temporal context aggregation through multi-order decomposition. For the spatial domain:
\begin{equation}
O_{2D} = \mathcal{S}_{2D}\big(\mathcal{F}_{\phi}(V),\; \{\mathcal{G}^{(\text{low})}_{\psi}, \mathcal{G}^{(\text{mid})}_{\psi}, \mathcal{G}^{(\text{high})}_{\psi}\}(V)\big),
\end{equation}
where a Feature Decomposition module ($\mathrm{FD}$) first suppresses trivial interactions via a $1\!\times\!1$ convolution followed by adaptive scaling with a learnable parameter initialized to zero; $\{\mathcal{G}^{(\cdot)}_{\psi}\}$ then computes multi-order context through depth-wise convolutions with kernel sizes $(5\!\times\!5, 5\!\times\!5, 7\!\times\!7)$ and dilations $(1, 2, 3)$ to capture low-order, middle-order, and high-order spatial interactions respectively; $\mathcal{F}_{\phi}$ performs aggregation gating via element-wise multiplication with SiLU activation; and $\mathcal{S}_{2D}$ fuses both branches.

For the temporal domain, MOTM processes each spatial position independently by reshaping $V$ into $(BHW, C, T)$, so that each spatial location $(h, w)$ is treated as an independent temporal sequence:
\begin{equation}
O_{1D} = \mathcal{S}_{1D}\big(\mathcal{F}_{\phi'}(V),\; \{\mathcal{G}^{(\text{low})}_{\psi'}, \mathcal{G}^{(\text{mid})}_{\psi'}, \mathcal{G}^{(\text{high})}_{\psi'}\}(V)\big),
\end{equation}
where a temporal decomposition first applies layer normalization over the channel dimension $C$, followed by adaptive readjustment via a learnable scaling factor $\gamma_t$ (initialized to zero) that suppresses trivial redundant inter-frame interactions; $\{\mathcal{G}^{(\cdot)}_{\psi'}\}$ computes multi-order temporal context through 1D depth-wise convolutions with kernel sizes $(5, 5, 7)$ and dilations $(1, 2, 3)$; $\mathcal{F}_{\phi'}$ performs a double gating aggregation where $\mathrm{Gate}_A$ operates on the normalized sequence as the aggregation branch and $\mathrm{Gate}_B$ operates on the concatenated multi-order features as the context branch; and $\mathcal{S}_{1D}$ fuses both branches. The resulting modulation signal $O_{1D} \in \mathbb{R}^{BT \times C \times H \times W}$ carries per-position temporal dynamics, preserving the full spatial structure without any broadcasting. The spatio-temporal fusion then follows:
\begin{equation}
O = O_{2D} \odot \big(1 + \alpha \cdot \tanh(O_{1D})\big),
\end{equation}
where $\alpha \in \mathbb{R}$ is a learnable scale initialized to zero for progressive training stability, and $\tanh$ bounds the modulation within $[-1, 1]$ to prevent gradient explosion. This ST-MOI design introduces a dual-level interaction between spatial and temporal dimensions, explicitly favoring middle-order interactions in both domains.

\subsection{VideoMogaNet Variants}
\label{sec:supp:variants}

The detailed architecture specifications of VideoMogaNet are shown in Table~\ref{tab:app_videomoga_architecture}, where an input video clip of $8 \times 224^2$ resolution (8 frames) is assumed for all architectures. Unlike image-based architectures, VideoMogaNet preserves the temporal dimension $T$ throughout all stages, applying spatial downsampling only. We adopt the same embedding dimension groups as~\cite{li2022moganet} while introducing temporal modeling capabilities through multi-order temporal modulation (MOTM) and spatio-temporal interaction (ST-MOI): \romannumeral1) VideoMogaNet-Tiny with embedding dimensions of $\{32, 64, 128, 256\}$ provides competitive efficiency for resource-constrained video understanding tasks; \romannumeral2) VideoMogaNet-Small adopts embedding dimensions of $\{64, 128, 320, 512\}$ for a favorable accuracy-efficiency trade-off; \romannumeral3) VideoMogaNet-Base with embedding dimensions of $\{64, 160, 320, 512\}$ targets balanced performance and efficiency for medium-scale video tasks; \romannumeral4) VideoMogaNet-Large with embedding dimensions of $\{64, 160, 320, 640\}$ is designed for large-scale video understanding applications. The GFLOPs are reported per view for video action recognition with 8-frame input clips at spatial resolution $224^2$, where, after all 4 stages, temporal average pooling is applied across frames followed by global spatial average pooling and a linear classifier. Each VideoMoga block contains spatial multi-order gated aggregation (SMixer), channel mixing (CMixer), multi-order temporal modulation (MOTM) and spatio-temporal interaction (ST-MOI) components, with the temporal dimension maintained constant across all stages.

\begin{table}[ht]
    \vspace{0.em}
    \setlength{\tabcolsep}{0.3mm}
    \centering
\resizebox{\columnwidth}{!}{
\begin{tabular}{ccc|cccc}
\toprule
\multicolumn{1}{c|}{Stage} & \multicolumn{1}{c|}{Output} & Layer & \multicolumn{4}{c}{VideoMogaNet} \\ \cline{4-7}
\multicolumn{1}{c|}{} & \multicolumn{1}{c|}{Size} & Settings & \multicolumn{1}{c|}{Tiny} & \multicolumn{1}{c|}{Small} & \multicolumn{1}{c|}{Base} & Large \\ \hline
\multicolumn{1}{c|}{\multirow{4}{*}{S1}} & \multicolumn{1}{c|}{\multirow{4}{*}{$T \times \frac{H}{4} \times \frac{W}{4}$}} & Stem & \multicolumn{4}{c}{\begin{tabular}[c]{@{}c@{}}$\rm{Conv}_{3\times 3},~\rm{stride}~2, C/2$ \\ $\rm{Conv}_{3\times 3},~\rm{stride}~2, C$\end{tabular}} \\ \cline{3-7}
\multicolumn{1}{c|}{} & \multicolumn{1}{c|}{} & Embed. Dim. & \multicolumn{1}{c|}{32} & \multicolumn{1}{c|}{64} & \multicolumn{1}{c|}{64} & 64 \\ \cline{3-7}
\multicolumn{1}{c|}{} & \multicolumn{1}{c|}{} & \# VideoMoga Block & \multicolumn{1}{c|}{3} & \multicolumn{1}{c|}{2} & \multicolumn{1}{c|}{4} & 4 \\ \cline{3-7}
\multicolumn{1}{c|}{} & \multicolumn{1}{c|}{} & MLP Ratio & \multicolumn{4}{c}{8} \\ \hline
\multicolumn{1}{c|}{\multirow{4}{*}{S2}} & \multicolumn{1}{c|}{\multirow{4}{*}{$T \times \frac{H}{8} \times \frac{W}{8}$}} & Stem & \multicolumn{4}{c}{$\rm{Conv}_{3\times 3},~\rm{stride}~2,~C_2$} \\ \cline{3-7}
\multicolumn{1}{c|}{} & \multicolumn{1}{c|}{} & Embed. Dim. & \multicolumn{1}{c|}{64} & \multicolumn{1}{c|}{128} & \multicolumn{1}{c|}{160} & 160 \\ \cline{3-7}
\multicolumn{1}{c|}{} & \multicolumn{1}{c|}{} & \# VideoMoga Block & \multicolumn{1}{c|}{3} & \multicolumn{1}{c|}{3} & \multicolumn{1}{c|}{6} & 6 \\ \cline{3-7}
\multicolumn{1}{c|}{} & \multicolumn{1}{c|}{} & MLP Ratio & \multicolumn{4}{c}{8} \\ \hline
\multicolumn{1}{c|}{\multirow{4}{*}{S3}} & \multicolumn{1}{c|}{\multirow{4}{*}{$T \times \frac{H}{16} \times \frac{W}{16}$}} & Stem & \multicolumn{4}{c}{$\rm{Conv}_{3\times 3},~\rm{stride}~2,~C_3$} \\ \cline{3-7}
\multicolumn{1}{c|}{} & \multicolumn{1}{c|}{} & Embed. Dim. & \multicolumn{1}{c|}{128} & \multicolumn{1}{c|}{320} & \multicolumn{1}{c|}{320} & 320 \\ \cline{3-7}
\multicolumn{1}{c|}{} & \multicolumn{1}{c|}{} & \# VideoMoga Block & \multicolumn{1}{c|}{12} & \multicolumn{1}{c|}{12} & \multicolumn{1}{c|}{22} & 44 \\ \cline{3-7}
\multicolumn{1}{c|}{} & \multicolumn{1}{c|}{} & MLP Ratio & \multicolumn{4}{c}{4} \\ \hline
\multicolumn{1}{c|}{\multirow{4}{*}{S4}} & \multicolumn{1}{c|}{\multirow{4}{*}{$T \times \frac{H}{32} \times \frac{W}{32}$}} & Stem & \multicolumn{4}{c}{$\rm{Conv}_{3\times 3},~\rm{stride}~2,~C_4$} \\ \cline{3-7}
\multicolumn{1}{c|}{} & \multicolumn{1}{c|}{} & Embed. Dim. & \multicolumn{1}{c|}{256} & \multicolumn{1}{c|}{512} & \multicolumn{1}{c|}{512} & 640 \\ \cline{3-7}
\multicolumn{1}{c|}{} & \multicolumn{1}{c|}{} & \# VideoMoga Block & \multicolumn{1}{c|}{2} & \multicolumn{1}{c|}{2} & \multicolumn{1}{c|}{3} & 4 \\ \cline{3-7}
\multicolumn{1}{c|}{} & \multicolumn{1}{c|}{} & MLP Ratio & \multicolumn{4}{c}{4} \\ \hline
\multicolumn{3}{c|}{Classifier} & \multicolumn{4}{c}{Temporal Avg. Pool, Global Spatial Avg. Pool, Linear} \\ \hline
\multicolumn{3}{c|}{Parameters (M)} & \multicolumn{1}{c|}{4.963} & \multicolumn{1}{c|}{24.872} & \multicolumn{1}{c|}{43.275} & 81.953 \\ \hline
\multicolumn{3}{c|}{GFLOPs per view} & \multicolumn{1}{c|}{8.806} & \multicolumn{1}{c|}{39.722} & \multicolumn{1}{c|}{79.398} & 127.297 \\
\bottomrule
\end{tabular}
    }
    \vspace{-0.5em}
    \caption{Architecture configurations of VideoMogaNet variants. Input resolution is $8 \times 224^2$ (8 frames). Unlike image-based models, the temporal dimension $T$ is preserved across all stages with spatial-only downsampling. GFLOPs are reported per view; effective inference cost on K400 is GFLOPs per view multiplied by 12.}
    \label{tab:app_videomoga_architecture}
    \vspace{-1.0em}
\end{table}

\noindent \textbf{Per-stage hyperparameters.}
All variants share a FFN expansion ratio of
$\{8, 8, 4, 4\}$ across stages and a tubelet size of $1$.
Tab.~\ref{tab:app_motm_hparams} reports the per-stage embedding
dimensions $C_i$ and block counts $N_i$ for each variant.
The channel growth ratio $C_{i+1}/C_i$ is variable across stages,
distinguishing VideoMogaNet from uniform-doubling backbones.

\begin{table}[h]
\centering
\footnotesize
\renewcommand{\arraystretch}{0.88}
\caption{Per-stage embedding dimensions $C_i$ and block counts $N_i$
for each VideoMogaNet variant. Growth ratio $C_{i+1}/C_i$ is shown
between stages.}
\label{tab:app_motm_hparams}
\resizebox{\linewidth}{!}{
\begin{tabular}{lcccccccc}
\toprule
Variant
  & $C_1$ & $C_2$ & $C_3$ & $C_4$
  & $N_1$ & $N_2$ & $N_3$ & $N_4$ \\
\midrule
Tiny  & 32 & 64  & 128 & 256 & 3 & 3 &  12 & 2 \\
Small & 64 & 128 & 320 & 512 & 2 & 3 &  12 & 2 \\
Base  & 64 & 160 & 320 & 512 & 4 & 6 &  22 & 3 \\
Large & 64 & 160 & 320 & 640 & 4 & 6 &  44 & 4 \\
\bottomrule
\end{tabular}}
\vspace{-.em}
\end{table}

\subsection{Pseudo-Code for MOTM and ST-MOI}
\label{sec:supp:pseudocode}

Algorithm~\ref{alg:motm} provides the pseudo-code for MOTM and ST-MOI.
The input shape is $(BT, C, H, W)$, where $B$, $T$, $C$, $H$, $W$
denote batch size, frames, channels, height, and width.
Channel groups follow the $1{:}3{:}4$ split with kernels
$k{=}(5,5,7)$ and dilations $d{=}(1,2,3)$.

\begin{algorithm}[t]
\caption{Pseudo-code for MOTM and ST-MOI}
\label{alg:motm}
\footnotesize
\begin{algorithmic}[1]

\Require Input $x \in \mathbb{R}^{BT \times C \times H \times W}$
\Require Post-CMixer features $\tilde{Y} \in \mathbb{R}^{BT \times C \times H \times W}$
\Statex Split: $C_0{:}C_1{:}C_2 = 1{:}3{:}4$,
kernels $k=(5,5,7)$, dilations $d=(1,2,3)$

\Function{init}{}
    \State $\mathrm{ln} \gets \mathrm{LayerNorm}(C)$
    \State $\gamma_t \gets \mathrm{Parameter}(\mathrm{zeros}(C))$
    \State $\mathrm{dw\_low} \gets \mathrm{Conv1d}(C_0,k=5,d=1,\mathrm{groups}=C_0)$
    \State $\mathrm{dw\_mid} \gets \mathrm{Conv1d}(C_1,k=5,d=2,\mathrm{groups}=C_1)$
    \State $\mathrm{dw\_high} \gets \mathrm{Conv1d}(C_2,k=7,d=3,\mathrm{groups}=C_2)$
    \State $\mathrm{gate\_A} \gets \mathrm{Conv1d}(C,k=1,\mathrm{groups}=C)$
    \State $\mathrm{gate\_B} \gets \mathrm{Conv1d}(C,k=1,\mathrm{groups}=C)$
    \State $\alpha \gets \mathrm{Parameter}(\mathrm{zeros}(1,C,1,1))$
\EndFunction

\Function{forward}{$x,\tilde{Y}$}

    \Statex \Comment{Dense Temporal Decomposition (Eqs.~\eqref{eq:td_ln}--\eqref{eq:td_gamma})}
    \State $y \gets \mathrm{reshape}(x,BHW,T,C)$
    \State $z \gets \mathrm{ln}(y)$
    \State $z \gets z+\gamma_t\odot(z-\mathrm{mean}(z,\mathrm{dim}=T))$

    \Statex \Comment{Parallel Multi-Order Extraction (Eqs.~\eqref{eq:motm_low}--\eqref{eq:motm_high})}
    \State $z \gets \mathrm{permute}(z,BHW,C,T)$
    \State $z_0,z_1,z_2 \gets \mathrm{split}(z,[C_0,C_1,C_2],\mathrm{dim}=1)$
    \State $z_0 \gets \mathrm{dw\_low}(z_0)$
    \State $z_1 \gets \mathrm{dw\_mid}(z_1)$
    \State $z_2 \gets \mathrm{dw\_high}(z_2)$

    \Statex \Comment{Adaptive Gating Aggregation (Eq.~\eqref{eq:motm_gate})}
    \State $\mathcal{F} \gets \mathrm{SiLU}(\mathrm{gate\_A}(z))$
    \State $\mathcal{G} \gets \mathrm{SiLU}(\mathrm{gate\_B}(\mathrm{cat}[z_0,z_1,z_2]))$
    \State $M \gets (\mathcal{F}\odot\mathcal{G}).\mathrm{reshape}(BT,C,H,W)$

    \Statex \Comment{ST-MOI Fusion (Eq.~\eqref{eq:stmoi})}
    \State $Z \gets \tilde{Y}\odot(1+\alpha\cdot\tanh(M))$

    \State \Return $Z$
\EndFunction

\end{algorithmic}
\end{algorithm}

\vspace{-.2em}
\section{Implementation Details}
\label{sec:supp:impl}

\noindent \textbf{Training Configurations}: Table~\ref{tab:training_config} summarizes the hyperparameters employed for all experiments across UCF-101, HMDB51, Kinetics-400 (K400), and Diving-48 (D-48) datasets. We adopt the AdamW optimizer with cosine annealing learning rate schedule. A consistent effective batch size of 256 is maintained across all experiments through gradient accumulation, following the formula: $\text{Batch}_{\text{total}} = \text{Batch}_{\text{per\_gpu}} \times \text{Num\_GPU} \times \text{Acc\_Steps}$. This strategy enables training on hardware configurations with limited GPU memory while preserving optimization dynamics. The learning rate is set to $8 \times 10^{-4}$ for all four benchmarks. Training epochs vary by dataset and model variant: on UCF-101, HMDB51, and D-48, Base trains for 120 epochs; on K400, Base/Large train for 30 epochs, Small for 50 epochs, and Tiny for 90 epochs. All models are initialized with~\cite{li2022moganet} weights pre-trained on ImageNet-1K. Training and fine-tuning are conducted on 4$\times$ NVIDIA A100-SXM4-40GB GPUs with an AMD EPYC 7413 24-Core Processor (2 sockets). Mixed-precision training with Apex AMP (O1, FP16) is employed for computational efficiency.

\begin{table}[h!]
\centering
\caption{Training configurations for VideoMogaNet across four benchmarks.}
\label{tab:training_config}
\small
\setlength{\tabcolsep}{3pt}
\renewcommand{\arraystretch}{0.95}
\vspace{-.4em}
\begin{tabular}{l|cccc}
\toprule
\textbf{Configuration} & \textbf{UCF-101} & \textbf{HMDB51} & \textbf{K400} & \textbf{D-48} \\
\midrule
optimizer & \multicolumn{4}{c}{\textit{AdamW}} \\
weight decay & \multicolumn{4}{c}{0.001} \\
LR schedule & \multicolumn{4}{c}{\textit{CosineAnnealing}} \\
learning rate & \multicolumn{4}{c}{$8 \times 10^{-4}$} \\
batch size & \multicolumn{4}{c}{256} \\
warmup epochs & \multicolumn{4}{c}{5} \\
gradient clipping & \multicolumn{4}{c}{1.0} \\
\midrule
\multicolumn{5}{l}{\textbf{Total Epochs (by variant)}} \\
\quad Tiny & $-$ & $-$ & 90 & $-$ \\
\quad Small & $-$ & $-$ & 50 & $-$ \\
\quad Base & 120 & 120 & 30 & 120 \\
\quad Large & $-$ & $-$ & 30 & $-$ \\
\midrule
drop path rate & \multicolumn{4}{c}{0.1} \\
\midrule
\multicolumn{5}{l}{\textbf{Data Augmentation}} \\
mixup & \multicolumn{4}{c}{0.8} \\
cutmix & \multicolumn{4}{c}{1.0} \\
mixup switch prob & \multicolumn{4}{c}{0.5} \\
label smoothing & \multicolumn{4}{c}{0.1} \\
color jitter & \multicolumn{4}{c}{0.8} \\
gray scale & \multicolumn{4}{c}{0.2} \\
flip ratio & \multicolumn{4}{c}{0.5} \\
\bottomrule
\end{tabular}
\normalsize
\vspace{-.4em}
\end{table}

\noindent \textbf{Evaluation Protocols}: We adopt multi-view evaluation strategies tailored to each dataset's characteristics, following standard video recognition practices~\cite{feichtenhofer2019slowfast}. For UCF-101, HMDB51, and Kinetics-400, we employ a $4\!\times\!3$ evaluation scheme (4 temporal clips $\times$ 3 spatial crops), resulting in 12 views per video. This configuration captures both temporal dynamics and spatial variations essential for the diverse action categories in these benchmarks. For Diving-48, we use a single-view $1\!\times\!1$ scheme (1 temporal clip $\times$ 1 spatial crop), consistent with standard evaluation protocols on this benchmark. Final predictions are obtained by averaging softmax probabilities across all views. All evaluations use spatial resolution $224^2$, with 8-frame input clips for UCF-101, HMDB-51, and K400, and 16-frame input clips for D-48. FLOPs measurements are conducted on an NVIDIA RTX A6000 GPU with an Intel Xeon Silver 4310 CPU @ 2.10GHz (2 sockets), reporting single-view inference costs for fair comparison with prior work.

\section{Multi-Order Interaction Analysis}
\label{sec:supp:empirical}

To validate that VideoMogaNet effectively promotes balanced multi-order
spatio-temporal interactions, we conduct a comprehensive game-theoretic
analysis on Kinetics-400. Throughout this section, we refer to
$\mathrm{SMixer}$, introduced in the main paper, as SMOGA (Spatial
Multi-Order Gated Aggregation), consistent with its original naming
in~\cite{li2022moganet}. This study comprises two complementary
components: (i) systematic comparison with baseline architectures
(ConvNeXt-T+Avg, MogaNet-S+Avg) to measure the gain brought by our
temporal contributions, and (ii) compositional validation decomposing
the respective contributions of MOTM and ST-MOI within the full
VideoMogaNet architecture.

\subsection{Game-Theoretic Interaction Framework}

\noindent\textbf{Revisiting multi-order interaction for images (2D).}
Following~\cite{deng2021discovering,zhang2020interpreting}, we interpret
learned representations through the lens of multi-order game-theoretic
interactions, which disentangle inter-variable communication effects into
diverse components of different interaction orders. The order here denotes
the \textit{scale of context} involved in the computation of the
game-theoretic interaction. Interactions of low order mainly reflect
\textbf{widely-shared local textures} and common visual concepts.
Middle-order interactions are primarily responsible for encoding
\textbf{discriminative high-level} representations.
High-order interactions are inclined to let DNNs memorize patterns of
\textbf{rare outliers} and large-scale shapes with intensive global
interactions, which can cause overfitting~\cite{deng2021discovering,
cheng2021game}. Consequently, the occurrence of excessively low- or
high-order interactions in a deep architecture is undesirable.

Formally, given an input image $x$ with a set of $n$ patches
$N = \{1,\dots,n\}$, the $m$-th order interaction $I^{(m)}(i,j)$
measures the average interaction effect between patches $i$ and $j$
across all contexts of size $m$:
\begin{equation}
I^{(m)}(i,j) = \mathbb{E}_{S \subseteq N \setminus \{i,j\},
|S|=m}[\Delta f(i,j,S)],
\label{eq:interaction}
\end{equation}
where $\Delta f(i,j,S) = f(S \cup \{i,j\}) - f(S \cup \{i\}) -
f(S \cup \{j\}) + f(S)$, and
$f(S) = \log \frac{P(\hat{y}=y \mid x_S)}{1 - P(\hat{y}=y \mid x_S)}$
denotes the log-odds score of the model on the masked input $x_S$,
where $y$ is the ground-truth label and patches in $N \setminus S$ are
replaced by a baseline value~\cite{ancona2019explaining}.
The relative interaction strength $J^{(m)}$ measures the distribution of
encoded interactions across orders:
\begin{equation}
J^{(m)} = \frac{\mathbb{E}_{x \in \Omega}\mathbb{E}_{i,j}
|I^{(m)}(i,j \mid x)|}{\mathbb{E}_{m'}\mathbb{E}_{x \in \Omega}
\mathbb{E}_{i,j}|I^{(m')}(i,j \mid x)|},
\label{eq:strength}
\end{equation}
where $\Omega$ is the set of all samples and $0 \leq m \leq n-2$.
Note that $J^{(m)}$ indicates a normalized distribution over interaction
orders (area under curve sums to one).

\noindent\textbf{Multi-order interaction for videos (3D).}
We propose a generalization of multi-order analysis to 3D spatio-temporal
grids. Given a video with $n = T \times H \times W$ patches where each
patch has 3D coordinates $(t, h, w)$, we define the unified 3D patch set
$N_{\text{3D}} = \{0, 1, \ldots, T \times H \times W - 1\}$ with 1D
indexing: $\text{idx} = t \cdot (H \times W) + h \cdot W + w$.
The spatio-temporal interaction becomes:
\begin{equation}
I^{(m)}_{\text{3D}}(i,j) = \mathbb{E}_{S \subseteq N_{\text{3D}}
\setminus \{i,j\}, |S|=m}[\Delta f_{\text{3D}}(i,j,S)],
\label{eq:interaction_3d}
\end{equation}
where $\Delta f_{\text{3D}}(i,j,S) = f_{\text{3D}}(S \cup \{i,j\}) -
f_{\text{3D}}(S \cup \{i\}) - f_{\text{3D}}(S \cup \{j\}) + f_{\text{3D}}(S)$
and $f_{\text{3D}}(\cdot)$ denotes the log-odds score on 3D-masked inputs.
The normalized interaction strength is defined as:
\begin{equation}
J^{(m)}_{\text{3D}} = \frac{\mathbb{E}_{x \in \Omega}\mathbb{E}_{i,j \in
N_{\text{3D}}}|I^{(m)}_{\text{3D}}(i,j \mid x)|}{\mathbb{E}_{m'}
\mathbb{E}_{x \in \Omega}\mathbb{E}_{i,j \in N_{\text{3D}}}
|I^{(m')}_{\text{3D}}(i,j \mid x)|}.
\label{eq:strength_3d}
\end{equation}

\noindent This unified formulation captures joint spatio-temporal
interactions: a pair $(i,j)$ can be purely spatial (same frame,
$t_i = t_j$), purely temporal (same spatial position), or mixed
(different frames and positions). This joint analysis aligns naturally
with VideoMogaNet's ST-MOI design:
\begin{equation}
Z = \tilde{Y} + \alpha \cdot \tilde{Y} \odot
\tanh\!\bigl(\mathrm{TMixer}(\tilde{Y})\bigr),
\end{equation}
where $\tilde{Y}$ denotes the spatially and channel-enriched features,
$\odot$ element-wise multiplication, and $\alpha \in \mathbb{R}$ a
learnable scale initialized to zero.

\subsection{Experimental Protocol}

\noindent\textbf{Dataset and architecture selection.}
We evaluate on the Kinetics-400 validation set.
Three architectures are compared:
(i)~ConvNeXt-T+Avg, a classical 2D ConvNet with temporal frame averaging;
(ii)~MogaNet-S+Avg, which incorporates $\mathrm{SMixer}$ (SMOGA) with
temporal frame averaging, providing a reference for architectures that
promote middle-order spatial interactions~\cite{li2022moganet};
and (iii)~VideoMogaNet-S, our full architecture introducing MOTM and
ST-MOI on top of SMOGA.
All models are initialized with ImageNet-1K pre-trained weights and
fine-tuned on K400.

\noindent\textbf{3D grid configuration.}
For each video, we define a spatio-temporal grid of
$T \times H \times W = 8 \times 14 \times 14 = 1{,}568$ patches.
Each 3D patch $(t, h, w)$ is mapped to a 1D index
$\text{idx} = t \cdot (H \times W) + h \cdot W + w \in [0, 1567]$,
following the unified indexing scheme of Eq.~\ref{eq:interaction_3d}.

\noindent\textbf{Pair and context sampling.}
We generate $N_{\text{pairs}} = 200$ patch pairs $(i,j)$ within a 3D
local neighborhood defined by $|t_i - t_j| \leq 1$,
$|h_i - h_j| \leq 1$, $|w_i - w_j| \leq 1$.
For each order ratio $r$ sampled from the set
\begin{equation*}
\mathcal{R} = \{0,\; 0.05,\; 0.1,\; 0.2,\; 0.3,\; 0.4,\; 0.5,\;
0.6,\; 0.7,\; 0.9,\; 1.0\},
\end{equation*}
we compute context size $m = \lfloor r \times (n-2) \rfloor$
and sample $N_{\text{ctx}} = 20$ random contexts
$S \subseteq N_{\text{3D}} \setminus \{i,j\}$ of cardinality $m$,
yielding $200 \times 20 = 4{,}000$ samples per order.

\noindent\textbf{Masked logit computation.}
For each video, pair $(i,j)$, and context $S$, we create four
3D-masked configurations replacing excluded patches with a uniform gray
baseline value of $0.5$.
The log-odds score $f_{\text{3D}}(\cdot) =
\log(p_{\text{gt}}/(1-p_{\text{gt}}))$ is computed for each
configuration, with probabilities clamped to $[10^{-7}, 1-10^{-7}]$.
The absolute strength $I^{(m)}_{\text{3D}}$ is averaged over all
contexts, pairs, and videos before normalization into
$J^{(m)}_{\text{3D}}$ following Eq.~\ref{eq:strength_3d}.

\noindent\textbf{Compositional ablation.}
To decompose the respective contributions of MOTM and ST-MOI, we
evaluate three configurations using the same protocol:
$\mathrm{SMixer}$ (SMOGA) alone (MogaNet-S+Avg, 24.809M\,/\,39.581G),
MOTM alone (0.063M\,/\,0.141G for VideoMogaNet-S), and ST-MOI
(VideoMogaNet-S, 24.872M\,/\,39.722G).

\vspace{-.4em}
\subsection{Visualization of Interaction Strength}

Interaction strength distributions $J^{(m)}_{\text{3D}}$ are visualized
in Fig.~\ref{fig:multi_order_and_st_ablation} (left) for the architecture
comparison and (right) for the compositional ablation.
These curves plot normalized interaction strength versus order ratio
($r=0$: low-order, $r=0.5$: middle-order, $r=1$: high-order).
Quantitative results are provided in
Tables~\ref{tab:interaction_arch} and~\ref{tab:interaction_ablation}.

\vspace{-.em}
\subsection{Quantitative Results}

\noindent\textbf{Architecture comparison.}
Table~\ref{tab:interaction_arch} presents the full
$J^{(m)}_{\text{3D}}$ distributions across the three architectures.
ConvNeXt-T+Avg exhibits a profile heavily concentrated at high-order
interactions ($J^{(1.0)} = 0.423$), reflecting its reliance on global
context under temporal averaging, with critically weak middle-order
encoding ($J^{(0.5)} = 0.035$). This distribution reflects the
inability to capture discriminative intermediate semantic compositions
essential for video action recognition.

MogaNet-S+Avg achieves a substantial middle-order improvement
($J^{(0.5)} = 0.051$, $+46\%$ vs ConvNeXt-T+Avg), validating the
effectiveness of $\mathrm{SMixer}$ (SMOGA) for encoding discriminative
intra-frame spatial interactions~\cite{li2022moganet}.
However, the absence of explicit temporal modeling limits inter-frame
coordination for actions spanning multiple frames.

VideoMogaNet-S achieves the strongest middle-order encoding
($J^{(0.5)} = 0.056$, $+59\%$ vs ConvNeXt-T+Avg,
$+9\%$ vs MogaNet-S+Avg).
This consistent progression ($0.035 \to 0.051 \to 0.056$) empirically
validates that ST-MOI coordination explicitly strengthens discriminative
middle-order interactions across both spatial and temporal dimensions,
confirming the central hypothesis of our work.

\begin{table}[t]
\centering
\caption{Multi-order interaction strength $J^{(m)}_{\text{3D}}$ on the
$8{\times}14{\times}14$ spatio-temporal grid (K400-trained).
VideoMogaNet-S achieves the strongest middle-order encoding at $m{=}0.5$,
with consistent improvement over both baselines.}
\label{tab:interaction_arch}
\setlength{\tabcolsep}{3pt}
\resizebox{\columnwidth}{!}{%
\begin{tabular}{c|ccc|ccc}
\toprule
\multirow{2}{*}{\textbf{Ratio}}
  & \multicolumn{3}{c|}{\textbf{$J^{(m)}_{\text{3D}}$}}
  & \multicolumn{3}{c}{\textbf{$I^{(m)}_{\text{abs}}$ ($\times 10^{-3}$)}} \\
& ConvNeXt-T+Avg & MogaNet-S+Avg & VideoMogaNet-S
& ConvNeXt-T+Avg & MogaNet-S+Avg & VideoMogaNet-S \\
\midrule
0.00 & 0.071 & \textbf{0.182} & 0.171 & 2.784 & 5.064 & 6.386 \\
0.05 & \textbf{0.097} & 0.108 & 0.109 & 3.804 & 2.993 & 4.087 \\
0.10 & 0.089 & 0.085 & 0.085 & 3.487 & 2.356 & 3.177 \\
0.20 & 0.057 & 0.066 & 0.066 & 2.238 & 1.823 & 2.473 \\
0.30 & 0.043 & 0.058 & 0.059 & 1.691 & 1.624 & 2.187 \\
0.40 & 0.037 & 0.054 & 0.057 & 1.447 & 1.499 & 2.123 \\
\rowcolor{gray!20}
\textbf{0.50} & 0.035 & 0.051 & \textbf{0.056}
              & 1.377 & 1.429 & 2.089 \\
0.60 & 0.034 & 0.054 & 0.055 & 1.347 & 1.491 & 2.066 \\
0.70 & 0.035 & 0.056 & 0.058 & 1.387 & 1.566 & 2.168 \\
0.90 & 0.077 & 0.080 & 0.082 & 3.027 & 2.222 & 3.079 \\
1.00 & \textbf{0.423} & 0.205 & 0.201 & 16.569 & 5.697 & 7.527 \\
\bottomrule
\end{tabular}%
}
\vspace{-1.em}
\end{table}

\noindent\textbf{Compositional ablation.}
Table~\ref{tab:interaction_ablation} decomposes the contributions of
MOTM and ST-MOI relative to the SMOGA spatial baseline.
$\mathrm{SMixer}$ (SMOGA) alone achieves effective middle-order spatial
encoding ($J^{(0.5)} = 0.051$), confirming that spatial multi-order
gated aggregation captures discriminative intra-frame part-whole
compositions, consistent with~\cite{li2022moganet} on images.

MOTM exhibits a markedly strong middle-order profile, peaking at
$m{=}0.3$ ($J^{(0.3)} = 0.154$) and remaining elevated at $m{=}0.5$
($J^{(0.5)} = 0.119$, $+133\%$ vs SMOGA), with the range
$r \in [0.3, 0.7]$ concentrating approximately $59\%$ of total
interaction strength.
This distribution confirms that our multi-order 1D depthwise
decomposition with progressive dilations explicitly targets
intermediate inter-frame interactions capturing action dynamics
spanning multiple frames, while attenuating both trivial
adjacent-frame redundancies (low-order) and potentially noisy
long-range correlations (high-order).
Critically, this temporal expressiveness is achieved at lightweight
cost: MOTM requires only $0.141$GFLOPs and $0.063$M parameters
($+0.25\%$ params, $+0.36\%$ GFLOPs relative to SMOGA).

ST-MOI coordinates SMOGA and MOTM through bounded multiplicative
modulation, achieving a balanced profile ($J^{(0.5)} = 0.056$) that
preserves SMOGA's spatial middle-order strength while integrating
MOTM's temporal dynamics.
The learnable scale $\alpha$ initialized to zero ensures training
begins from the spatial-only identity baseline and progressively
activates temporal modulation, while $\tanh$ bounding prevents
gradient instabilities.
These results validate the complementarity of SMOGA and MOTM: the
former encodes discriminative spatial structures within each frame,
while the latter adaptively modulates these structures according to
multi-order temporal context, and their coordination via ST-MOI
promotes the emergence of robust spatio-temporal middle-order
interactions.

\begin{table}[t]
\centering
\caption{Compositional ablation on the $8{\times}14{\times}14$ grid
(K400-trained). MOTM achieves a strong middle-order peak
($J^{(0.3)}{=}0.154$, $J^{(0.5)}{=}0.119$, $+133\%$ vs SMOGA) at
lightweight cost ($0.141$GFLOPs for VideoMogaNet-S).
ST-MOI coordination produces a balanced profile that preserves spatial
middle-order strength while integrating temporal dynamics.}
\label{tab:interaction_ablation}
\setlength{\tabcolsep}{3pt}
\resizebox{\columnwidth}{!}{%
\begin{tabular}{c|ccc|ccc}
\toprule
\multirow{2}{*}{\textbf{Ratio}}
  & \multicolumn{3}{c|}{\textbf{$J^{(m)}_{\text{3D}}$}}
  & \multicolumn{3}{c}{\textbf{$I^{(m)}_{\text{abs}}$ ($\times 10^{-3}$)}} \\
& SMOGA & MOTM & ST-MOI & SMOGA & MOTM & ST-MOI \\
\midrule
0.00 & \textbf{0.182} & 0.064 & 0.171 & 5.064 & 2.525 & 6.386 \\
0.05 & 0.108 & 0.050 & 0.109 & 2.993 & 1.999 & 4.087 \\
0.10 & 0.085 & 0.049 & 0.085 & 2.356 & 1.940 & 3.177 \\
0.20 & 0.066 & 0.117 & 0.066 & 1.823 & 4.639 & 2.473 \\
0.30 & 0.058 & \textbf{0.154} & 0.059 & 1.624 & 6.107 & 2.187 \\
0.40 & 0.054 & 0.138 & 0.057 & 1.499 & 5.487 & 2.123 \\
\rowcolor{gray!20}
\textbf{0.50} & 0.051 & \textbf{0.119} & \textbf{0.056}
              & 1.429 & 4.731 & 2.089 \\
0.60 & 0.054 & 0.104 & 0.055 & 1.491 & 4.127 & 2.066 \\
0.70 & 0.056 & 0.072 & 0.058 & 1.566 & 2.856 & 2.168 \\
0.90 & 0.080 & 0.036 & 0.082 & 2.222 & 1.439 & 3.079 \\
1.00 & 0.205 & 0.097 & 0.201 & 5.697 & 3.847 & 7.527 \\
\midrule
\multicolumn{4}{l|}{\textit{Params (M) / FLOPs (G)}} &
  \multicolumn{1}{c|}{24.809 / 39.581} &
  \multicolumn{1}{c|}{0.063 / 0.141} &
  \multicolumn{1}{c}{24.872 / 39.722} \\
\bottomrule
\end{tabular}
}
\vspace{-1.2em}
\end{table}

\noindent\textbf{Discussion.}
This game-theoretic analysis provides three levels of empirical
validation.
First, at the architecture level, VideoMogaNet-S achieves optimal
middle-order spatio-temporal modeling ($J^{(0.5)} = 0.056$), with a
consistent progression over both baselines ($+59\%$ vs
ConvNeXt-T+Avg, $+9\%$ vs MogaNet-S+Avg), demonstrating that ST-MOI
coordination effectively promotes discriminative middle-order
interactions that passive temporal averaging cannot capture.
Second, at the module level, the compositional ablation confirms the
synergistic complementarity of MOTM and SMOGA: SMOGA efficiently
encodes spatial middle-order interactions intra-frame
($J^{(0.5)} = 0.051$), while MOTM explicitly forces temporal
middle-order encoding inter-frame with a strong peak at $m{=}0.3$
($J^{(0.3)} = 0.154$, $J^{(0.5)} = 0.119$, $+133\%$ vs SMOGA) at
lightweight cost ($0.141$GFLOPs for VideoMogaNet-S, compared to the
tens of GFLOPs required by attention-based temporal encoders in
architectures such as VideoSwin~\cite{liu2022video} or
TimeSformer~\cite{bertasius2021space}), and ST-MOI coordinates both
into a balanced and robust spatio-temporal profile.
Third, this analysis is conducted under a statistically robust
protocol ($N_{\text{pairs}}=200$, $N_{\text{ctx}}=20$,
$4{,}000$ samples per order), confirming that explicit multi-order
decomposition combined with bounded multiplicative modulation suffices
to capture discriminative spatio-temporal interactions without costly
global attention mechanisms.
These properties are reflected in the empirical benchmarks:
VideoMogaNet achieves $97.1\%$ on UCF-101, $74.7\%$ on HMDB51,
$81.3\%$ on Kinetics-400, and $90.30\%$ on Diving-48
(see main paper, Sec.~\ref{subsec:comparison}).


\section{Computational Complexity Analysis}
\label{sec:supp:complexity}

We analyze the computational complexity of a VideoMoga block for an input
$V \in \mathbb{R}^{C \times T \times H \times W}$, where $C$, $T$, $H$, and $W$
denote the number of channels, temporal length, height, and width respectively,
and $B$ denotes the batch size.
The analysis mirrors the three-component decomposition of the VideoMoga block
(Eqs.~(\ref{eq:smixer})--(\ref{eq:stmixer}) of the main paper): the spatial
backbone built upon~\cite{li2022moganet}, our temporal module MOTM
(Sec.~\ref{subsec:motm}), and our spatio-temporal fusion ST-MOI
(Sec.~\ref{subsec:st_interaction}).
The spatial cost is derived first, as it provides the reference denominator
against which the overhead of MOTM and ST-MOI is measured.

\subsection{Spatial Backbone: \texorpdfstring{$\mathrm{SMixer}$}{SMixer} (SMOGA) and Channel Aggregation}
\label{sec:supp:flops_spatial}

The $\mathrm{SMixer}(\cdot)$ and $\mathrm{CMixer}(\cdot)$ modules
(Eqs.~(\ref{eq:smixer})--(\ref{eq:cmixer}) of the main paper) build on the
spatial design of~\cite{li2022moganet}, extended here to operate on each
of the $T$ frames independently as $(BT, C, H, W)$ tensors within our
spatio-temporal architecture.

\noindent\textbf{SMOGA cost.}
$\mathrm{SMixer}(\cdot)$ first applies $\mathrm{DW}_{5\times5,\,d=1}$
to all $C$ channels, then factorizes the output into three groups
$Y_l \in \mathbb{R}^{C_l \times HW}$, $Y_m \in \mathbb{R}^{C_m \times HW}$,
$Y_h \in \mathbb{R}^{C_h \times HW}$ with $C_l + C_m + C_h = C$, and applies
$\mathrm{DW}_{5\times5,\,d=2}$ and $\mathrm{DW}_{7\times7,\,d=3}$ to
$Y_m$ and $Y_h$ respectively.
The cost of the three depthwise convolutions per frame is:
\begin{align}
    \mathrm{FLOPs}(\mathrm{DW}_{2D})
    &= 2HWC \cdot 25 + \tfrac{3}{4}HWC \cdot 25 \nonumber \\ +  HWC \cdot 49 
    &= 117.75\,HWC,
    \label{eq:supp:dw2d}
\end{align}
where the three terms correspond to $\mathrm{DW}_{5\times5,\,d=1}$ over
all $C$ channels, $\mathrm{DW}_{5\times5,\,d=2}$ over $C_m = 3C/8$ channels,
and $\mathrm{DW}_{7\times7,\,d=3}$ over $C_h = C/2$ channels.
The four $1\!\times\!1$ pointwise projections of the feature decomposition
and gated aggregation steps contribute $6HWC^2$ FLOPs per frame, yielding:
\begin{equation}
    \mathrm{FLOPs}(\mathrm{SMOGA}_{\mathrm{frame}}) = HWC\!\left(6C + 117.75\right).
    \label{eq:supp:smoga}
\end{equation}

\noindent\textbf{Channel Aggregation cost.}
$\mathrm{CMixer}(\cdot)$ implements an inverted bottleneck
MLP~\cite{cvpr2018mobilenetv2} with expansion ratio $r{=}8$ at stages 1--2
and $r{=}4$ at stages 3--4, a depthwise $3{\times}3$ convolution, and a channel
decomposition step, contributing $HWC(16C + 72)$ FLOPs per frame.

\vspace{1.em}
\noindent\textbf{\textbf{Total spatial cost.}}
Summing $\mathrm{SMixer}(\cdot)$ and $\mathrm{CMixer}(\cdot)$ over $T$ frames,
we define the spatial cost as:
\begin{equation}
    F_{\mathrm{Sp}} \triangleq \mathrm{FLOPs}(\mathrm{Spatial}) = T \cdot HWC\!\left(22C + 189.75\right),
    \label{eq:supp:spatial}
\end{equation}
which scales as $\mathcal{O}(T \cdot H \cdot W \cdot C^2)$ and is the
dominant cost of a VideoMoga block, equivalent to extending the spatial
design of~\cite{li2022moganet} to $T$ frames independently.

\subsection{Multi-Order Temporal Modulation (MOTM)}
\label{sec:supp:flops_motm}

MOTM instantiates $\mathrm{TMixer}(\cdot)$ (Eq.~(\ref{eq:tmixer}) of the main
paper). As described in Sec.~\ref{subsec:motm}, it operates by reshaping the
input from $\mathbb{R}^{BT \times C \times H \times W}$ to
$\mathbb{R}^{BHW \times T \times C}$, so that each spatial position $(h, w)$
is processed as an independent temporal sequence of length $T$, before
permuting to $\mathbb{R}^{BHW \times C \times T}$ for the convolution steps.
This reshape is parameter-free and introduces no FLOPs.

\vspace{1.em}
\noindent\textbf{Depthwise 1D convolutions.}
As defined in Eqs.~(\ref{eq:motm_low})--(\ref{eq:motm_high}) of the main
paper, the $C$ channels are partitioned into three groups
$z_0 \in \mathbb{R}^{BHW \times C_0 \times T}$,
$z_1 \in \mathbb{R}^{BHW \times C_1 \times T}$,
$z_2 \in \mathbb{R}^{BHW \times C_2 \times T}$,
where $C_0 + C_1 + C_2 = C$, with
$C_0 = C/2$, $C_1 = C/8$, and $C_2 = 3C/8$.
Each group is processed by a dedicated $\mathrm{DWConv1D}$ with
$\mathrm{groups} = C_i$, so that each channel is filtered independently along
$T$ with no cross-channel projection.
For a depthwise 1D convolution over $C_i$ channels with kernel size $k_i$,
the FLOPs count is $2 \cdot BHW \cdot C_i \cdot T \cdot k_i$.
Summing over the three branches:
\begin{align}
    \mathrm{FLOPs}(\mathrm{DW}_{1D})
    &= 2\,BHWT\!\left(
        \tfrac{C}{2} \cdot 5 + \tfrac{C}{8} \cdot 5 + \tfrac{3C}{8} \cdot 7
    \right) \nonumber\\
    &= 11.5\,BHWCT,
    \label{eq:supp:dw1d}
\end{align}
where the three terms correspond to $\mathrm{DWConv1D}_{k=5,\,d=1}$ over
$C_0 = C/2$ channels, $\mathrm{DWConv1D}_{k=5,\,d=2}$ over $C_1 = C/8$
channels, and $\mathrm{DWConv1D}_{k=7,\,d=3}$ over $C_2 = 3C/8$ channels.
Since all three convolutions are strictly depthwise, no $C^2$ cross-channel
term arises from multi-order temporal extraction.

\vspace{1.em}
\noindent\textbf{\textbf{Gating aggregation.}}
As defined in Eq.~(\ref{eq:motm_gate}) of the main paper,
$\mathrm{Gate}_A$ and $\mathrm{Gate}_B$ are two grouped $1{\times}1$
$\mathrm{Conv1d}$ layers with $\mathrm{groups} = C$, each contributing
$2\,BHWCT$ FLOPs, for a gating total of $4\,BHWCT$.

\vspace{1.em}
\noindent\textbf{Total MOTM cost.}
Combining Eqs.~(\ref{eq:supp:dw1d}) and the gating cost, we define:
\begin{equation}
\label{eq:supp:motm}
\begin{split}
F_{\mathrm{MOTM}}
&\triangleq \mathrm{FLOPs}(\mathrm{MOTM}) \\
&= 11.5\,BHWCT + 4\,BHWCT \\
&= 15.5 \cdot B \cdot H \cdot W \cdot C \cdot T.
\end{split}
\end{equation}
which scales as $\mathcal{O}(B \cdot T \cdot H \cdot W \cdot C)$, one order in $C$
below the spatial baseline of Eq.~(\ref{eq:supp:spatial}).
Note that Eq.~(\ref{eq:supp:motm}) excludes the LayerNorm
(Eq.~(\ref{eq:td_ln}) of the main paper) and the temporal decomposition
parameter $\gamma_t \in \mathbb{R}^C$ (Eq.~(\ref{eq:td_gamma}) of the main
paper), which contribute lower-order terms in $C$ and are accounted for
in the discussion below.

\vspace{1.em}
\noindent\textbf{Overhead relative to the spatial baseline.}
The relative overhead of MOTM with respect to
Eq.~(\ref{eq:supp:spatial}) is
\begin{align}
    \varepsilon_{\mathrm{MOTM}}
    &= \frac{\mathrm{FLOPs}(\mathrm{MOTM})}{\mathrm{FLOPs}(\mathrm{Spatial})}
    = \frac{15.5\,B}{22C + 189.75} \nonumber\\
    &\xrightarrow{C \to \infty} \frac{15.5\,B}{22C},
    \label{eq:supp:eps}
\end{align}
which scales as $\mathcal{O}(1/C)$. Consequently, the relative temporal
cost decreases monotonically with increasing channel dimension and becomes
negligible for larger models. For VideoMogaNet-S ($B{=}1$), the
FLOPs-weighted global overhead predicted by Eq.~(\ref{eq:supp:eps}) is
$0.31\%$, closely matching the measured $+0.36\%$ reported in the main
paper. The small discrepancy is fully explained by the $\mathrm{LN}$ and
$\gamma_t$ terms omitted from Eq.~(\ref{eq:supp:motm}), confirming the
tightness of the analytical bound.

\subsection{Spatio-Temporal Multi-Order Interaction (ST-MOI)}
\label{sec:supp:flops_stmoi}
ST-MOI instantiates $\mathrm{STMixer}(\cdot)$ (Eq.~(\ref{eq:stmixer}) of the
main paper). As defined in Eq.~(\ref{eq:stmoi}) of the main paper, the
dual-level spatio-temporal interaction is:
\begin{equation}
    Z = \tilde{Y} + \alpha \cdot \tilde{Y}
        \odot \tanh\!\bigl(\mathrm{TMixer}(\tilde{Y})\bigr),
    \label{eq:supp:stmoi}
\end{equation}
where $\tilde{Y} \in \mathbb{R}^{BT \times C \times H \times W}$ denotes the
output of $\mathrm{CMixer}(\cdot)$ (Eq.~(\ref{eq:cmixer}) of the main paper),
$\odot$ denotes element-wise multiplication, $\alpha \in \mathbb{R}$ is a
learnable scaling factor initialized to zero, and $\tanh(\cdot)$ bounds the
modulation within $[-1, 1]$ for training stability.
Setting $M = \mathrm{TMixer}(\tilde{Y}) \in \mathbb{R}^{BT \times C \times H \times W}$,
the four operations in Eq.~(\ref{eq:supp:stmoi}), namely $\tanh(M)$,
$\tilde{Y} \odot \tanh(M)$, $\alpha \cdot (\cdot)$, and the outer addition,
each operate element-wise over $BTHWC$ entries, yielding:
\begin{equation}
    F_{\mathrm{ST\text{-}MOI}} \triangleq \mathrm{FLOPs}(\mathrm{ST\text{-}MOI})
    = 4 \cdot B \cdot T \cdot C \cdot H \cdot W,
    \label{eq:supp:stmoi_flops}
\end{equation}
which scales as $\mathcal{O}(B \cdot T \cdot H \cdot W \cdot C)$, with zero
additional parameters beyond $\alpha \in \mathbb{R}$.

\subsection{Total Cost and Comparison}
\label{sec:supp:flops_total}
\begin{table}[t]
\centering
\caption{Asymptotic complexity comparison. $n_t$: temporal length;
$n_h \times n_w$: spatial resolution; $d$: channel dimension.
GFLOPs are reported for single-view inference (${\times}12$ for
UCF-101, HMDB-51, and K400, single-view for D-48), consistent with
Sec.~\ref{subsec:setup} of the main paper.
$\dagger$: spatial design extended from~\cite{li2022moganet} to the video
domain.}
\label{tab:supp:complexity}
\setlength{\tabcolsep}{4pt}
\renewcommand{\arraystretch}{1.15}
\resizebox{\columnwidth}{!}{%
\begin{tabular}{llc}
\toprule
Architecture & Complexity & Temporal scaling \\
\midrule
ViViT~\cite{arnab2021vivit}
  & $\mathcal{O}\!\left((n_h n_w)^2 d + n_t^2 d\right)$
  & Quadratic in $n_t$ \\
VideoMamba~\cite{park2024videomamba}
  & $\mathcal{O}(n_h n_w n_t d)$
  & Linear \\
\midrule
\textbf{VideoMogaNet} (total)
  & $\mathcal{O}(n_t n_h n_w d^2)$
  & Linear in $n_t$ \\
\quad Spatial (SMOGA+CA)$^\dagger$
  & $\mathcal{O}(n_t n_h n_w d^2)$
  & Inherited $d^2$ term \\
\quad MOTM (\textbf{ours})
  & $\mathcal{O}(n_t n_h n_w d)$
  & No $d^2$ term (depthwise only) \\
\quad ST-MOI (\textbf{ours})
  & $\mathcal{O}(n_t n_h n_w d)$
  & No $d^2$ term (element-wise only) \\
\bottomrule
\end{tabular}%
}
\vspace{-1.4em}
\end{table}

Using the spatial cost $F_{\mathrm{Sp}}$ (Eq.~(\ref{eq:supp:spatial})) and the
temporal contributions $F_{\mathrm{MOTM}}$ (Eq.~(\ref{eq:supp:motm})) and
$F_{\mathrm{ST\text{-}MOI}}$ (Eq.~(\ref{eq:supp:stmoi})), the total cost per
VideoMoga block is:
\begin{equation}
    \mathrm{FLOPs}(\mathrm{VideoMogaBlock})
    = F_{\mathrm{Sp}} + F_{\mathrm{MOTM}} + F_{\mathrm{ST\text{-}MOI}},
    \label{eq:supp:total}
\end{equation}
where $F_{\mathrm{Sp}}$ is the inherited spatial term and
$F_{\mathrm{MOTM}} + F_{\mathrm{ST\text{-}MOI}}$ are our two contributions,
which is dominated by the spatial term for all practical values of $C$
and reduces to $\mathrm{FLOPs}(\mathrm{VideoMogaBlock}) \approx 22\,THWC^2$.
For a complete network with $N_i$ VideoMoga blocks at stage $i$ and
channel dimension $C_i$ at spatial resolution $H_i \times W_i$:
\vspace{-.4em}
\begin{align}
    \mathrm{FLOPs}(\mathrm{VideoMogaNet})
    &\approx 22 \cdot T \cdot \sum_{i=1}^{4} N_i \cdot H_i \cdot W_i \cdot C_i^2.
    \label{eq:supp:network}
\end{align}

\noindent\textbf{Discussion.}
MOTM and ST-MOI together account for the entirety of VideoMogaNet's temporal
modeling capability at a combined cost of $19.5\,BHWCT$ FLOPs, one order in
$C$ below the inherited spatial baseline of Eq.~(\ref{eq:supp:spatial}).
This efficiency follows directly from two structural choices: MOTM eliminates
all $\mathcal{O}(C^2)$ cross-channel projections by operating exclusively with
depthwise 1D convolutions throughout
Eqs.~(\ref{eq:motm_low})--(\ref{eq:motm_gate}) of the main paper, while
ST-MOI (Eq.~(\ref{eq:stmoi})) realizes dual-level
spatio-temporal fusion through element-wise operations only, adding no
convolutions and no $C^2$ term. By contrast, temporal attention in
VideoSwin~\cite{liu2022video} and TimeSformer~\cite{bertasius2021space} requires
tens of GFLOPs for temporal encoding alone, and a direct 1D transposition of
the spatial MOGA module (MOGA-1D baseline) costs $21.1$GFLOPs versus
$0.282$GFLOPs for MOTM, representing
${\sim}70{\times}$ reduction at a marginal $0.3\%$ Top-1 difference on K400.
Eq.~(\ref{eq:supp:eps}) further shows
that this advantage compounds in larger variants, where MOTM's overhead
decreases as $\mathcal{O}(1/C)$ while its multi-order expressiveness is
preserved, as validated by the game-theoretic interaction analysis of
Sec.~\ref{sec:supp:empirical}.

\vspace{1.em}
\noindent\textbf{Complexity comparison.}
Tab.~\ref{tab:supp:complexity} summarizes the asymptotic complexity of
VideoMogaNet alongside representative video architectures, using unified
notation $n_t = T$, $n_h \times n_w = H \times W$, and $d = C$.
ViViT~\cite{arnab2021vivit} with factorized attention incurs quadratic cost
in both spatial resolution and temporal length,
$\mathcal{O}((n_h n_w)^2 d + n_t^2 d)$, as noted in Sec.~\ref{sec:intro}
of the main paper.
VideoMamba~\cite{park2024videomamba} achieves linear complexity
$\mathcal{O}(n_h n_w n_t d)$ through selective state-space models.
VideoMogaNet inherits a $d^2$ term from its spatial backbone, a necessary
cost for multi-order cross-channel interaction, while the two contributions
of this work, MOTM and ST-MOI, are both strictly linear in all dimensions.

\vspace{-.4em}
\section{Extended Experiments}
\label{sec:supp:extended}

\noindent \textbf{Model Scalability.}
Table~\ref{tab:scalability} reports results across four model variants
on Kinetics-400, all using ImageNet-1K pre-training, 8 frames, and
$4{\times}3$ views. VideoMogaNet exhibits consistent scalability: Tiny
($76.7\%$) $\to$ Small ($79.6\%$, $+2.9\%$) $\to$ Base ($80.6\%$,
$+1.0\%$) $\to$ Large ($81.3\%$, $+0.7\%$), from a lightweight
configuration (Tiny: $8.806$G / $4.963$M) to a large-scale one (Large:
$127.297$G / $81.953$M), without requiring the costly global attention
mechanisms typical of heavy transformer-based models.

\begin{table}[t]
\centering
\caption{Scalability of VideoMogaNet variants on Kinetics-400. All models
use ImageNet-1K pre-training, 8 frames, and $4{\times}3$ views.
Unlike Tab.~3 of the main paper, GFLOPs here are per-view costs;
multiply by $4{\times}3$ to obtain effective inference cost.}
\label{tab:scalability}
\setlength{\tabcolsep}{4pt}
\renewcommand{\arraystretch}{1.15}
\resizebox{\columnwidth}{!}{
\begin{tabular}{l|ccc|ccc}
\toprule
\textbf{Model} & \textbf{Pretrain} & \textbf{Frames} & \textbf{Views}
  & \textbf{GFLOPs per view} & \textbf{Params (M)} & \textbf{Top-1 (\%)} \\
\midrule
VideoMogaNet-T & IN-1K & 8 & $4{\times}3$ &  8.806 &  4.963 & 76.7 \\
VideoMogaNet-S & IN-1K & 8 & $4{\times}3$ & 39.722 & 24.872 & 79.6 \\
VideoMogaNet-B & IN-1K & 8 & $4{\times}3$ & 79.398 & 43.275 & 80.6 \\
VideoMogaNet-L & IN-1K & 8 & $4{\times}3$ & 127.297 & 81.953 & \textbf{81.3} \\
\bottomrule
\end{tabular}
}
\vspace{-1.em}
\end{table}

\vspace{-.4em}
\section{Additional Information}
\label{sec:supp:additional}

\noindent\textbf{Dataset Licensing.}
All datasets used in this work are standard academic benchmarks with
publicly available licenses.
Kinetics-400\footnote{\url{https://www.deepmind.com/open-source/kinetics}}
and
HMDB51\footnote{\url{https://serre-lab.clps.brown.edu/resource/hmdb-a-large-human-motion-database}}
are licensed under CC BY 4.0\footnote{\url{https://creativecommons.org/licenses/by/4.0}},
UCF-101\footnote{\url{https://www.crcv.ucf.edu/data/UCF101.php}} is made
publicly available by the Center for Research in Computer Vision for
academic research purposes,
and
Diving-48
is released for academic research use.
All experiments comply with the respective dataset terms and conditions
for academic research.

\vspace{1.em}
\noindent\textbf{Broader Impact.}
VideoMogaNet addresses key practical challenges in large-scale video
recognition deployment through efficiency-focused design. The MOTM
module adds $0.282$GFLOPs and $0.063$M parameters to the spatial
backbone, a significant reduction compared to attention-based
approaches requiring tens of GFLOPs for temporal encoding alone.
Scalability from lightweight (Tiny: $8.806$GFLOPs / $4.963$M) to large-scale
(Large: $127.297$GFLOPs / $81.953$M) variants enables flexible deployment
across resource-constrained edge devices and high-performance servers.
Training on public academic benchmarks (UCF-101, HMDB51, Kinetics-400,
Diving-48) ensures reproducibility and transparency. To foster broader
adoption, we will release pre-trained VideoMogaNet weights on all four
benchmarks.

\vspace{1.em}
\noindent\textbf{Limitations.}
VideoMogaNet is designed and evaluated for single-person, clip-level
action recognition. Extending the approach to multi-person action
recognition, where multiple actors may perform distinct or interacting
actions within the same clip, would require additional architectural
adaptations to disentangle and attribute spatio-temporal interactions
to each individual. This constitutes a natural direction for future
work that may broaden the scope of VideoMogaNet while preserving the
computational efficiency that forms a core aspect of our contribution.
\clearpage

\begin{figure*}[p]
\centering
\includegraphics[
    width=1.0\textwidth,
    height=0.82\textheight,
    keepaspectratio
]{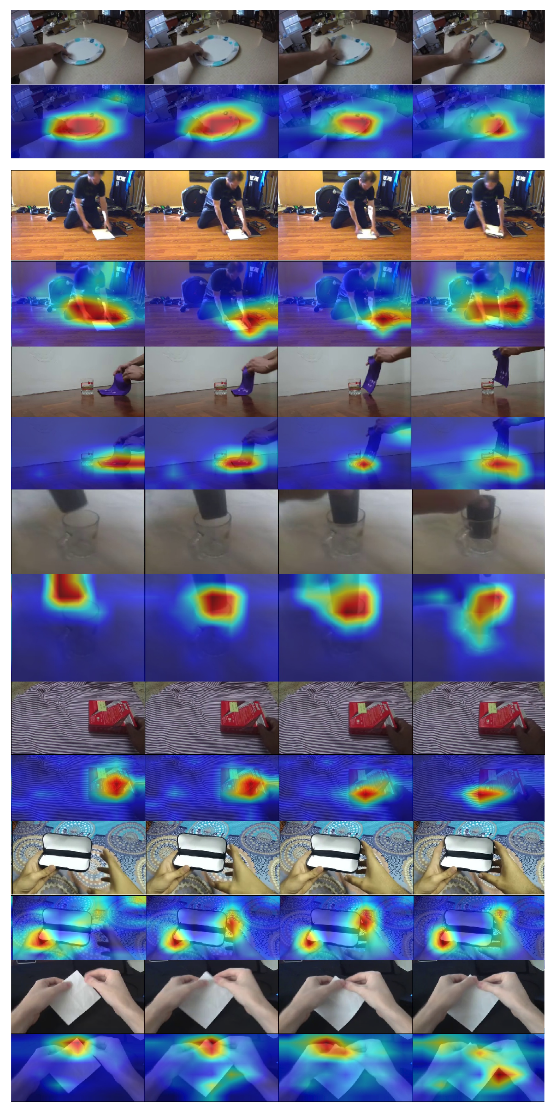}
\caption{\textbf{Grad-CAM visualization.} VideoMogaNet's spatio-temporal multi-order interaction (ST-MOI) effectively localizes and tracks action-relevant regions across temporal sequences. The architecture demonstrates precise focus on discriminative spatial regions where fine-grained object transformations occur (e.g., hand movements, object manipulations), while maintaining temporal consistency through multi-order temporal modulation (MOTM). Highlighted regions show the model's ability to capture both spatial part-whole compositions via $\mathrm{SMixer}$ (SMOGA) and temporal dynamics via MOTM, validating the synergistic coordination between spatial and temporal modalities for fine-grained action understanding.}
\label{fig:gradcam_ssv2}
\end{figure*}

\begin{figure*}[p]
\centering
\includegraphics[
    width=1.0\textwidth,
    height=0.82\textheight,
    keepaspectratio
]{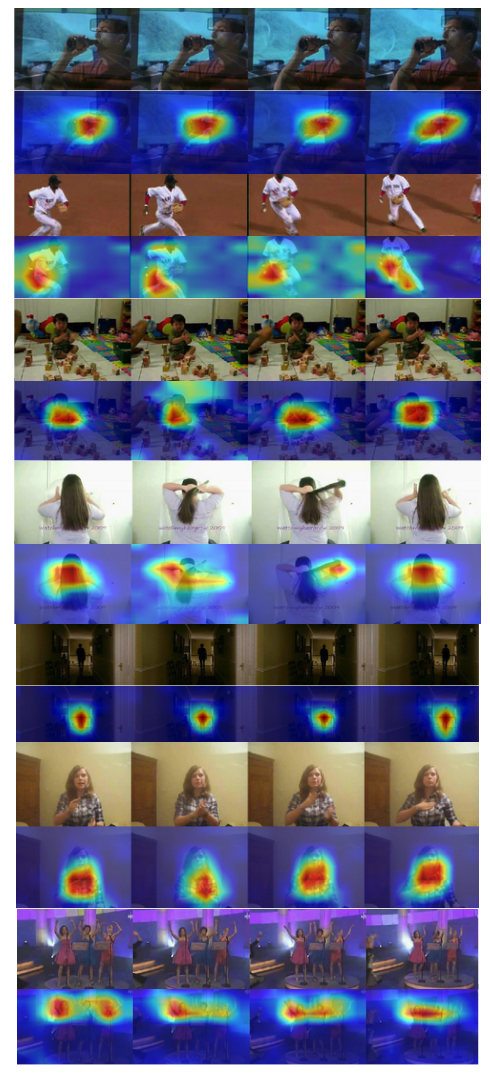}
\caption{\textbf{Grad-CAM visualization on HMDB51.} VideoMogaNet's ST-MOI architecture accurately localizes and tracks action-performing regions throughout video sequences. Attention maps demonstrate effective spatio-temporal coordination: $\mathrm{SMixer}$ (SMOGA) captures discriminative spatial structures (e.g., body parts, motion trajectories), while MOTM modulates these representations according to temporal context, enabling robust tracking of action dynamics across frames. The visualizations confirm that bounded multiplicative modulation ($1 + \alpha \cdot \tanh(\mathcal{M}_\theta)$) preserves spatial middle-order strength while amplifying temporal signals, resulting in consistent focus on semantically relevant regions for general action recognition.}
\label{fig:gradcam_hmdb51}
\end{figure*}


\end{document}